%% file: main_arxiv.tex
\documentclass[twocolumn,pagebackref,breaklinks,colorlinks,allcolors=iccvblue]{fairmeta}

\definecolor{iccvblue}{rgb}{0.21,0.49,0.74}

\usepackage{url}
\usepackage{makecell}

\usepackage{epsfig}
\usepackage{graphicx}
\usepackage{amsmath}
\usepackage{amssymb}
\usepackage{fontawesome5}
\usepackage{graphicx}
\usepackage{amsmath}
\usepackage{amssymb}
\usepackage{booktabs}
\usepackage[ruled,vlined]{algorithm2e}
\usepackage{algorithmic}
\usepackage{colortbl} 
\usepackage{booktabs} 
\usepackage{multirow, booktabs, graphicx, xcolor, colortbl, arydshln, tikz}
\usetikzlibrary{tikzmark}
\usepackage{makecell}
\usepackage{siunitx} 
\usepackage{mdframed}
\usepackage{tablefootnote} 

\newcolumntype{g}{>{\columncolor{gray!30}}c}

\usepackage{tcolorbox}
\tcbuselibrary{most}

\definecolor{citecolor}{HTML}{0071bc}
\hypersetup{
    colorlinks=true,%
    citecolor=citecolor,%
    filecolor=citecolor,%
    linkcolor=citecolor,%
    urlcolor=citecolor
}

\tcbuselibrary{skins}

\definecolor{userbg}{RGB}{245, 245, 245}
\definecolor{userborder}{RGB}{210, 229, 255}
\definecolor{userfont}{RGB}{0, 0, 0}

\tcbset{
  enhanced,
  boxrule=1pt,
  colback=userbg,
  colframe=userborder,
  fonttitle=\bfseries,
  coltitle=userfont,
  boxsep=5pt,
  arc=5pt,
  outer arc=5pt,
  left=10pt,
  right=10pt,
  top=2pt,
  bottom=2pt,
  attach boxed title to top left={yshift=-0.25\baselineskip-1mm,xshift=2mm},
  boxed title style={
    colback=userborder,
    sharp corners,
    rounded corners=northeast,
    arc=3pt,
    outer arc=3pt,
    boxrule=0pt,
    boxsep=2pt,
  },
}
\usepackage{pgf}
\usepackage{adjustbox}

\usepackage{multirow}
\usepackage{colortbl} 
\usepackage{booktabs}
\usepackage{tcolorbox} 
\usepackage{enumitem}
\usepackage{graphicx}
\usepackage{multirow}
\usepackage{array}
\definecolor{listcolor}{RGB}{50,120,230}
\usepackage{arydshln}
\usepackage{url}
\usepackage{caption}

\newcommand{\ourssl}{Web-SSL}
\newcommand{\ourdino}{Web-DINO}
\newcommand{\ourmae}{Web-MAE}
\newcommand{\ourdata}{MC-2B}

\newcounter{researchquestion}

\newcommand{\researchquestion}[2][]{%
  \vspace{0.8em}
  \refstepcounter{researchquestion}
  \begin{tcolorbox}[
    enhanced,
    colback=blue!5,
    colframe=blue!70!black,
    fonttitle={\fontsize{10.5pt}{12.8pt}\selectfont\bfseries\color{blue!20!black}},  
    title=Question \theresearchquestion,
    toprule=1.5pt,
    bottomrule=0.8pt,
    leftrule=0.8pt,
    rightrule=0.8pt,
    left=6pt,
    right=6pt,
    top=6pt,
    bottom=6pt,
    boxsep=3pt
  ]
  \normalsize #2
  \end{tcolorbox}
  \ifx\\#1\\\else\label{rq:#1}\fi
  \vspace{0.5em}
}

\title{Scaling Language-Free Visual Representation Learning}

\author[1,*]{David Fan}
\author[1,2,*]{Shengbang Tong}
\author[1,2]{Jiachen Zhu}
\author[1]{Koustuv Sinha}
\author[1,3]{Zhuang Liu}
\author[1]{Xinlei Chen}
\author[1]{Michael Rabbat}
\author[1]{Nicolas Ballas}
\author[1,2]{Yann LeCun}
\author[1,\dagger]{Amir Bar}
\author[2,\dagger]{Saining Xie}

\affiliation[1]{FAIR, Meta}
\affiliation[2]{New York University}
\affiliation[3]{Princeton University}
\contribution[*]{equal contribution}
\contribution[\dagger]{equal advising}

\abstract{
Visual Self-Supervised Learning (SSL) currently underperforms Contrastive Language-Image Pretraining (CLIP) in multimodal settings such as Visual Question Answering (VQA). This multimodal gap is often attributed to the semantics introduced by language supervision, even though visual SSL and CLIP models are often trained on different data. In this work, we ask the question: ``Do visual self-supervised approaches lag behind CLIP due to the lack of language supervision, or differences in the training data?''
We study this question by training both visual SSL and CLIP models on the same MetaCLIP data,
and leveraging VQA as a diverse testbed for vision encoders. In this controlled setup, visual SSL models scale better than CLIP models in terms of data and model capacity, and visual SSL performance does not saturate even after scaling up to 7B parameters. Consequently, we observe visual SSL methods achieve CLIP-level performance on a wide range of VQA and classic vision benchmarks.
These findings demonstrate that pure visual SSL can match language-supervised visual pretraining at scale, opening new opportunities for vision-centric representation learning.

}

\date{April 1, 2025}

\project{\href{https://davidfan.io/webssl/}{\sffamily \fontsize{8.8pt}{11pt}\selectfont \texttt{https://davidfan.io/webssl/}}}

\begin{document}
\maketitle

\section{Introduction}
\label{sec:intro}

Visual representation learning has evolved along two distinct paths with different training approaches. Language-supervised methods such as Contrastive Language-Image Pretraining (CLIP) \citep{radford2021learning, zhai2023sigmoid} use paired image-text data to learn representations that are enriched with linguistic semantics. Self-Supervised Learning (SSL) methods~\citep{zhang2016colorful,chen2020simple,he2022masked,lecun2022path,oquab2023dinov2} learn from images alone, without language.

\begin{figure}[t!]
    \centering
    \includegraphics[width=\linewidth]{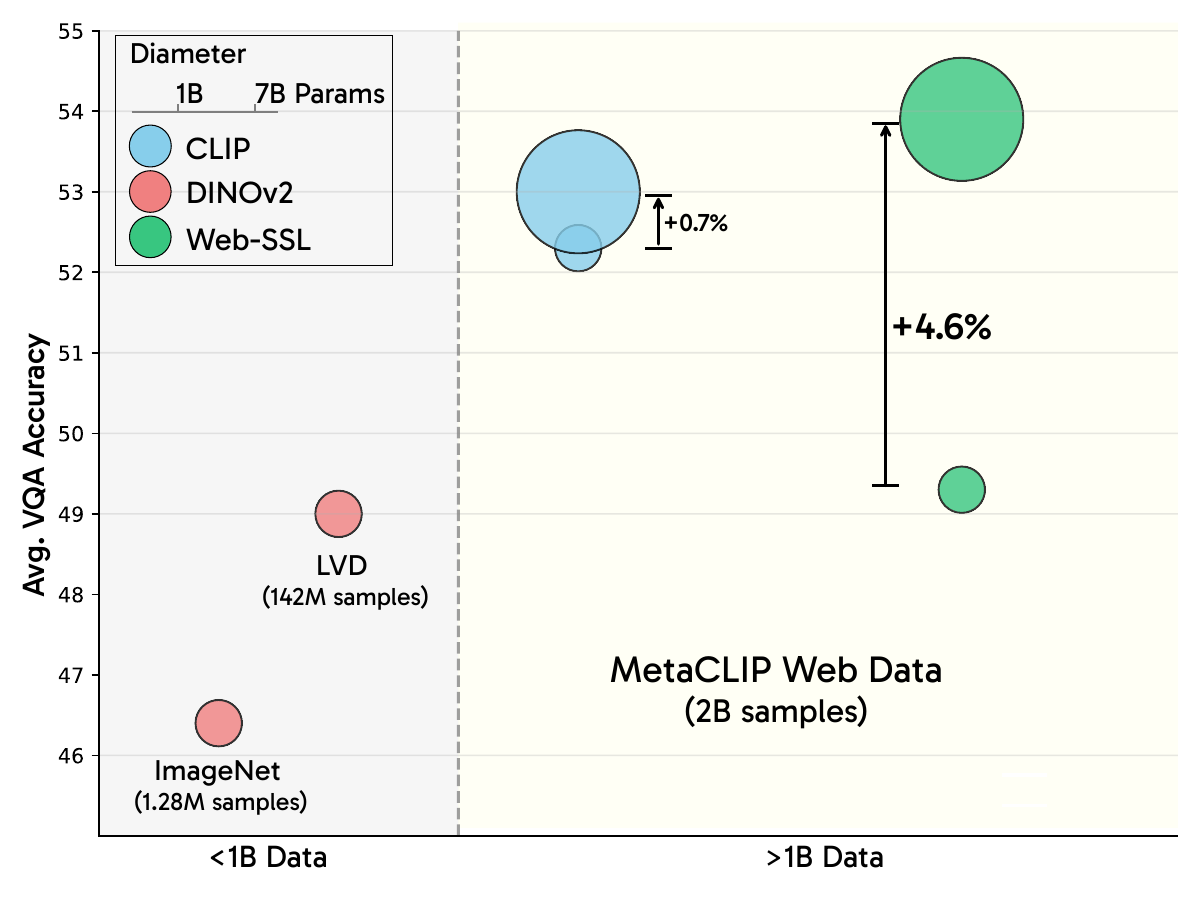}

    \caption{
    We compare the scaling behavior of visual SSL and CLIP on 16 VQA tasks from the Cambrian-1 suite under different data and model size regimes. Prior visual SSL methods achieved strong performance on classic vision tasks, but have underperformed as encoders for multimodal instruction-tuned VQA tasks. Our results show that with appropriate scaling of models and data, visual SSL can match the performance of language-supervised models across all evaluated domains---even OCR \& Chart.}
   
    \label{fig:teaser}
    \vspace{-0.5cm}
\end{figure}

Despite SSL models outperforming language-supervised models on classic vision tasks such as classification and segmentation \citep{oquab2023dinov2}, they are less commonly adopted in recent multimodal large language models (MLLMs) \citep{liu2023visual, liu2023improved, agrawal2024pixtral, tong2024cambrian, beyer2024paligemma, li2024llava, llama3modelcard}. This difference in adoption is partially due to a performance gap in visual question answering (see \cref{fig:teaser}), particularly for OCR \& Chart interpretation tasks~\citep{tong2024cambrian, shi2024eagle}.

\begin{figure*}[t!]
    \centering
    \includegraphics[width=\linewidth]{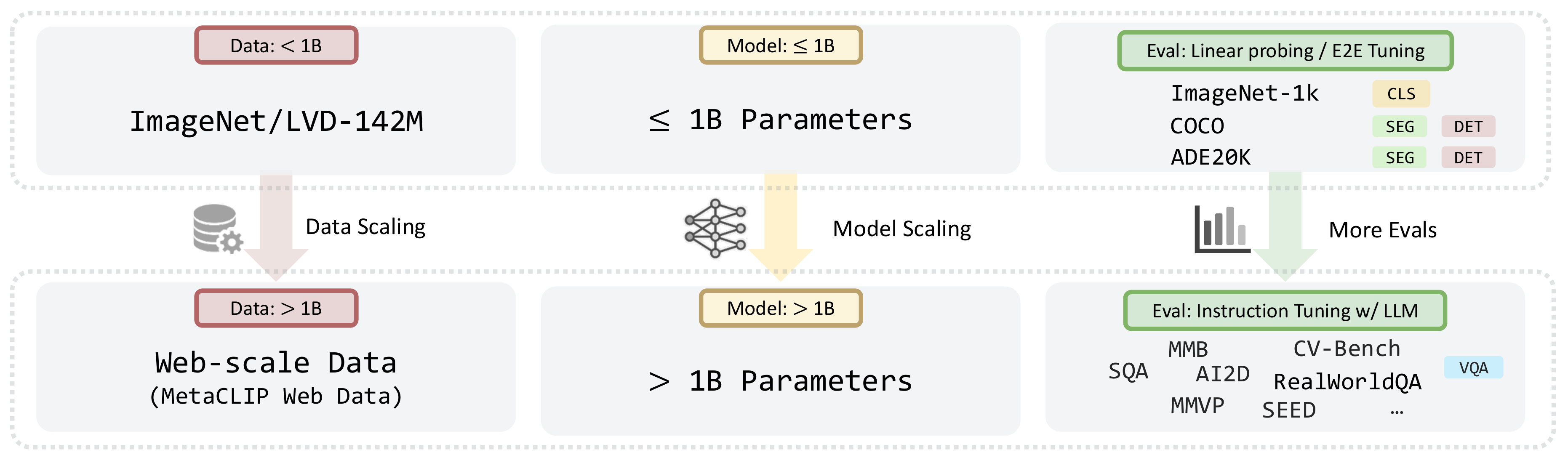}
  
    \caption{
    \textbf{Visual SSL 2.0 changes.} In this work, we adopt three improvements to the visual SSL pipeline:  1) Training on billion-scale web data, curated through the MetaCLIP pipeline, to move beyond ``conventional'' datasets; 2) Scaling model architecture from sub-billion parameter models to models exceeding 1 billion parameters; and 3) Incorporating VQA as a complementary evaluation protocol to comprehensively assess visual features. These changes enable us to study visual SSL at a larger scale and observe scaling trends previously unobserved in smaller-scale experiments.}

    \label{fig:pin demo}
    \vspace{-0.2cm}
\end{figure*}

Beyond methodology differences, these approaches have also been separated by data scale and distribution (\cref{fig:teaser}). CLIP models typically train on billion-scale image-text pairs from the web~\citep{schuhmann2022laion, chen2022pali, xu2023demystifying}, while SSL methods use million-scale datasets such as ImageNet~\citep{deng2009imagenet} or hundred-million scale data with ImageNet-like distributions~\citep{ridnik2021imagenet, oquab2023dinov2}.

In this work, we investigate a fundamental question: \textit{Is language supervision necessary to pretrain visual representations for multimodal modeling?} Rather than seeking to replace language-supervised approaches, we aim to understand the intrinsic capabilities and limitations of visual self-supervision at scale for multimodal applications. To conduct a fair comparison, we train SSL models on the same billion-scale web data used for state-of-the-art CLIP models---specifically the MetaCLIP dataset \citep{xu2023demystifying}. This approach controls for data distribution differences when comparing visual SSL and CLIP.

For evaluation, we primarily use visual question answering (VQA) as a framework to evaluate SSL models across a diverse set of capabilities at scale. VQA evaluation suites span vision-centric, visual reasoning, and OCR \& Chart tasks, and have been shown to be a more diverse testbed for assessing vision encoders~\citep{tschannen2024image, wan2024locca, fini2024multimodal,tong2024cambrian}, reflecting the broader perception challenges
found in real-world distributions. 
We adopt the evaluation suite proposed in Cambrian-1~\citep{tong2024cambrian}, which evaluates performance across 16 tasks spanning 4 distinct categories of VQA: General, Knowledge, OCR \& Chart, and Vision-Centric. 

We train \ourssl{}, a family of visual SSL models ranging from 1 to 7 billion parameters, using the above setting for direct and controlled comparison to CLIP. As a result of our empirical study, we contribute several insights: 

\begin{itemize}

\item Visual SSL can match and even surpass language-supervised methods for visual pretraining, on a wide range of VQA tasks---even on language-related tasks such as OCR \& Chart understanding (\cref{fig:scale vision}).

\item Visual SSL scales well with respect to model capacity (\cref{fig:scale vision}) and data (\cref{fig:scaling data}), indicating that SSL has significant untapped potential.

\item Visual SSL can maintain competitive traditional vision performance on classification and segmentation, even while improving at VQA (\cref{fig:classic benchmarks}).

\item Training on a higher ratio of images containing text is especially effective for improving OCR \& Chart performance (Question~\ref{rq:probe into text}). Exploring data composition is a promising direction.
\end{itemize}
This work serves as a proof of concept that offers a compelling vision-centric alternative to the recent CLIP-dominated trend, and opens new opportunities for future research.
We plan to open-source our \ourssl{} vision models, and we hope to inspire the broader community to unlock the full potential of visual SSL in the multimodal era.

\section{From Visual SSL 1.0 to 2.0}

\label{sec:methods}
In this section, we describe our experimental setup, which extends previous SSL works by (1) scaling dataset size to billion-scale images (\cref{subsec:beyond imagenet}), (2) scaling model size beyond 1B parameters (\cref{subsec:scale model sizes}), and (3) evaluating vision models using open-ended VQA tasks (\cref{subsec:MLLM as evaluation}), in addition to classic vision benchmarks such as ImageNet-1k~\citep{deng2009imagenet} and ADE20k~\citep{zhou2019semantic}.

\subsection{Beyond ImageNet Pretraining} \label{subsec:beyond imagenet}
To study whether visual SSL can match the performance of CLIP, we start by adopting the same data that drove CLIP's success.
We thus leverage the MetaCLIP dataset~\citep{xu2023demystifying, xu2024altogether}, which has enabled the most successful open-source reproduction of CLIP to-date.\footnote{The data used to train the original CLIP is closed-source.} We use 2 billion samples from MetaCLIP, which we refer to as \ourdata{}. We train SSL methods on only the images, and CLIP on the image-text pairs.

This controls for data distribution and size as confounding variables, and enables a fairer comparison of the pretraining methods themselves,
while ensuring sufficient data diversity and scale. 

\subsection{Scaling Up Vision Models to Billion Scale} \label{subsec:scale model sizes}
We can also increase model size. Inspired by advancements in scaling language models~\citep{brown2020language, kaplan2020scaling, OpenAI2022ChatGPT}, we train Vision Transformers (ViTs) with 1B, 2B, 3B, 5B, and 7B parameters, on only the images from \ourdata{}, to study the properties of larger-scale visual SSL models trained on web-scale data. We adapt ViT-g from \citet{oquab2023dinov2} as ViT-1B, and define new configurations for ViT-2B to 7B (\cref{tab:vit config}); see \cref{appdendix:implementation details} for model details.

\begin{table}[h]
    \centering
    \small
    \vspace{-0.1cm}
    \setlength{\tabcolsep}{3pt}
    \begin{tabular}{lccccc}
    \toprule
    Model & Width & Depth & Heads & MLP \\
    \midrule
    ViT-1B & 1536 & 40 & 24 & 6144  \\
    ViT-2B & 2688 & 24 & 21 & 10752  \\
    ViT-3B & 3072 & 26 & 24 & 12288  \\
    ViT-5B & 3584 & 32 & 28 & 14336  \\
    ViT-7B & 4096 & 32 & 32 & 16384  \\
    \bottomrule
    \end{tabular}
    \vspace{-0.1cm}
    \caption{
    \textbf{Model architecture details.} For consistency, we denote ViT-g from \citet{oquab2023dinov2} as ViT-1B.}
    \label{tab:vit config}
    \vspace{-0.3cm}
\end{table}
\subsection{Multimodal LLMs as an Evaluation Protocol} \label{subsec:MLLM as evaluation}
\label{sec:using MLLM as evaluator}

In addition to conventional evaluation protocols, such as ImageNet-1k linear probe, we also evaluate our vision encoders using VQA, a flexible and robust evaluation protocol that reflects the diversity of real-world perceptual challenges~\citep{tschannen2024image, tong2024cambrian}, as shown in~\cref{fig:pin demo}.

Here, we study all vision encoders using the same controlled setting to ensure fair comparison. Specifically, we use the same two-stage visual instruction tuning procedure and data as Cambrian-1~\citep{tong2024cambrian}. First, a lightweight MLP adapter is added to project the vision encoder features into the same dimensionality as the LLM, and only this MLP adapter is trained. In the second stage, both the MLP adapter and LLM are finetuned. To enable controlled comparison, the vision encoder remains frozen in both stages, and all experiments use the same training recipe as well as Llama-3 8B Instruct~\citep{touvron2023llama} backbone. We provide detailed training datasets and hyperparameters in \cref{appdendix:implementation details}.

We then report results on the Cambrian-1~\citep{tong2024cambrian} evaluation suite, which is comprised of 16 VQA benchmarks spanning four established domains: General, Knowledge, OCR \& Chart, and Vision-Centric. The average VQA performance is the average of the four subcategories. Each subcategory has 4 benchmarks and is equally weighted.

\section{Scaling Visual SSL}
\label{sec:exp}

\begin{figure*}[t!]
    \centering
    \includegraphics[width=1.0\linewidth]{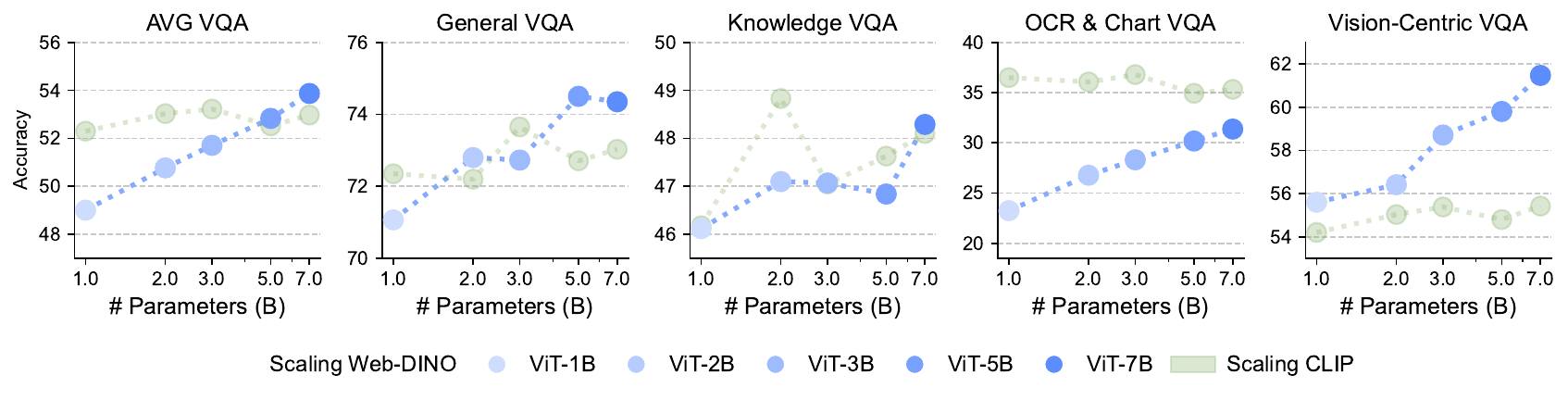}
\caption{
\textbf{Scaling behavior of \ourdino{} and CLIP ViTs trained on \ourdata{}.} The x-axis shows model sizes from 1B to 7B parameters on a log scale. We observe novel ``scaling behavior'' with \ourdino{} models across all categories, with particularly pronounced improvements in the OCR \& Chart and Vision-Centric domains as model size increases. In contrast, CLIP models demonstrate limited scaling benefits, with performance saturating at moderate model sizes. The two model families exhibit complementary strengths: CLIP models excel at OCR \& Chart VQA, and \ourdino{} models are superior at Vision-Centric VQA, while remaining competitive in all other categories.}
    \label{fig:scale vision}
    \vspace{-0.1cm}
\end{figure*}

In this section, we explore the scaling behavior of visual SSL models with respect to both model and data size, as a result of training on only images from \ourdata{}. We focus on DINOv2~\citep{oquab2023dinov2} as the visual SSL method in this section, and discuss MAE~\citep{he2022masked} in \cref{sec:analysis}.

In \cref{subsec:scale vit}, we increase model size from 1B to 7B while keeping the training data fixed at 2 billion \ourdata{} images---unless otherwise denoted. We use the off-shelf training code and recipe for each method, and do not change the recipe for different model sizes in order to control for confounding variables. In \cref{subsec:scale data}, we shift our focus to scaling total data seen for a fixed model size, and analyze how performance evolves as the number of images seen during training increases from 1 billion to 8 billion.

\subsection{Scaling Model}
\label{subsec:scale vit}

The intention of scaling model size is both to find the ceiling of visual SSL under this new data regime, and to identify any unique behavior that emerges in larger models.

We thus pretrain DINOv2 ViT models, ranging from 1B to 7B parameters, using 2 billion unlabeled images at 224$\times$224 resolution from \ourdata{}---\emph{without} high-resolution adaptation~\citep{oquab2023dinov2}---to ensure fair comparison with CLIP. We refer to these models as \ourdino{} throughout the paper. For a controlled comparison, we also train CLIP models of the same sizes on the same data. 

We evaluate each model with VQA and present the results in \cref{fig:scale vision}. We will first discuss the overall performance trend and then turn to specific category performance. To the best of our knowledge, this is the first instance of a vision encoder trained purely with visual self-supervision achieving performance parity with language-supervised encoders on VQA---even in the OCR \& Chart category, which is traditionally considered to be highly text-dependent.

\vspace{1mm}
\noindent\textbf{Performance trend.} We compare the performance trend as model capacity increases in \cref{fig:scale vision}. \ourdino{}'s Average, OCR \& Chart, and Vision-Centric VQA performance improves nearly log-linearly with increasing model size, while General and Knowledge improve to a smaller degree. In contrast, CLIP's performance in all VQA categories largely saturates after 3B parameters. This suggests that while smaller CLIP models may be more data-efficient, this advantage largely dissipates for larger CLIP models. The continual improvement from increasing \ourdino{} model capacity also suggests that visual SSL benefits from larger model capacity, and that scaling visual SSL past 7B parameters is a promising direction.

\vspace{1mm}
\noindent\textbf{Category-specific performance.} In terms of category-specific performance, DINO also increasingly outperforms CLIP on Vision-Centric VQA and largely closes the gap with CLIP on OCR \& Chart and Average VQA (\cref{fig:scale vision}), as model size increases. At 5B parameters and above, DINO can exceed the Average VQA performance of CLIP, despite being trained solely on images and without language supervision. 
These results suggest that vision-only models, when trained on CLIP-distribution images, can develop strong visual features that are comparable to those of language-supervised vision encoders.

\subsection{Scaling Examples Seen}
\label{subsec:scale data}
\begin{figure*}[t]
    \centering
    \includegraphics[width=\linewidth]{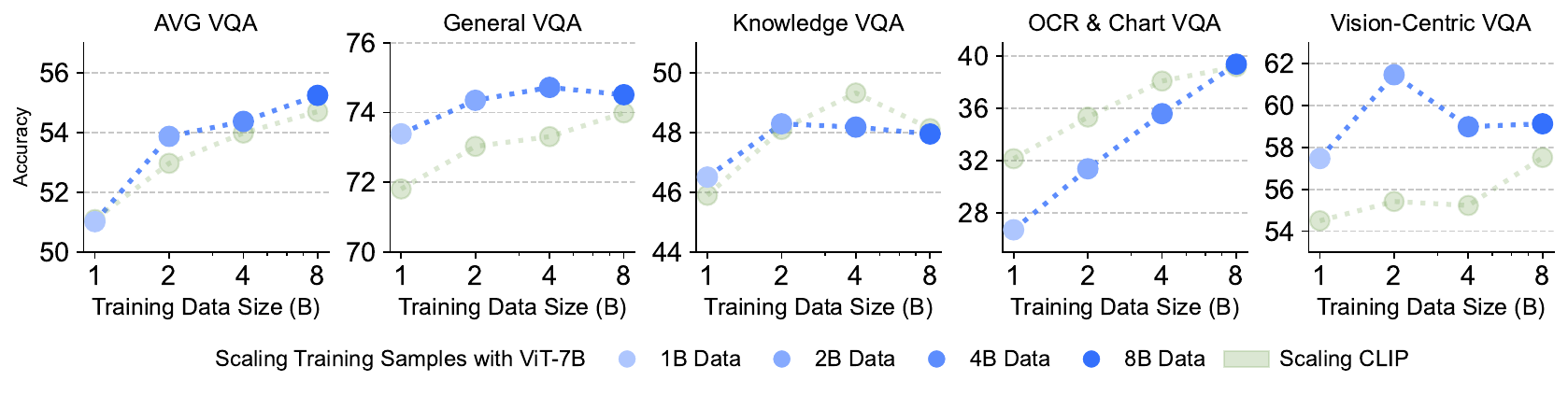}

\caption{
\textbf{Scaling up examples seen when training \ourdino{}-7B.} Performance across different VQA categories as training data increases from 1B to 8B images. While General and Vision-Centric tasks show diminishing returns after 2B images, OCR \& Chart tasks demonstrate continued improvement, contributing to steady gains in average performance. Further, \ourdino{} consistently outperforms same-size (ViT-7B) CLIP models with different training samples seen. The x-axis plots training data size on a log-scale.}
    \label{fig:scaling data}
    \vspace{-0.2cm}
\end{figure*}

Previously, we focused on single-epoch training, where each of the 2B unique images in \ourdata{} is seen only once. Here, we investigate the impact of increasing the number of examples seen by training \ourdino{} ViT-7B on data ranging from 1 billion to 8 billion images from \ourdata{}. 

As shown in \cref{fig:scaling data}, General and Knowledge VQA performance improves incrementally with more examples seen, saturating at 4B and 2B examples respectively. Vision-Centric VQA performance improves sharply from 1B to 2B examples, and saturates beyond 2B examples. In contrast, OCR \& Chart is the only category that shows consistent improvement with more examples seen. This suggests that as the model sees more data, it learns a representation that is increasingly well-suited for text-related tasks, yet without marked degradation on other capabilities.

Furthermore, when compared to a CLIP model of the same size (ViT-7B), \ourdino{} consistently outperforms CLIP on average VQA performance given the same number of samples seen (\cref{fig:scaling data}). Notably, after seeing 8B samples, \ourdino{} closes the performance gap with the CLIP model on OCR \& Chart VQA tasks. This provides further evidence suggesting that visual SSL models have the potential to scale better than language-supervised models.

Collectively, the results in \cref{fig:scale vision} and \ref{fig:scaling data} indicate that as model size and examples seen increase, visual SSL learns features that are increasingly effective for VQA in general, but especially on OCR \& Chart. Our results suggest that CLIP-based models do not hold an absolute advantage compared to visual SSL. In \cref{sec:analysis}, we delve deeper into the underlying mechanisms driving this trend.

\section{Scaling Analysis and Findings}
\label{sec:analysis}
In \cref{sec:exp}, we demonstrated that visual SSL models scale well with model size and training set size. These observations raise further questions about the generality and implications of these phenomena. To deepen our understanding, we investigate five key aspects, including whether scaling behavior extends to other vision-only models (Question~\ref{rq:generalize to other ssl}), if SSL models also exhibit scaling behavior on smaller and more conventional data (Question~\ref{rq:small data}), and whether SSL can retain competitive performance on classic vision tasks (Question~\ref{rq:classic vision}). Additionally, we explore why scaling particularly enhances OCR \& Chart performance (Question~\ref{rq:probe into text}), and highlight emergent properties that arise via scaling visual SSL (Question~\ref{rq:properties}). In this section, we provide a detailed analysis of these findings.

\begin{figure*}[t!]
    \centering
    \includegraphics[width=\linewidth]{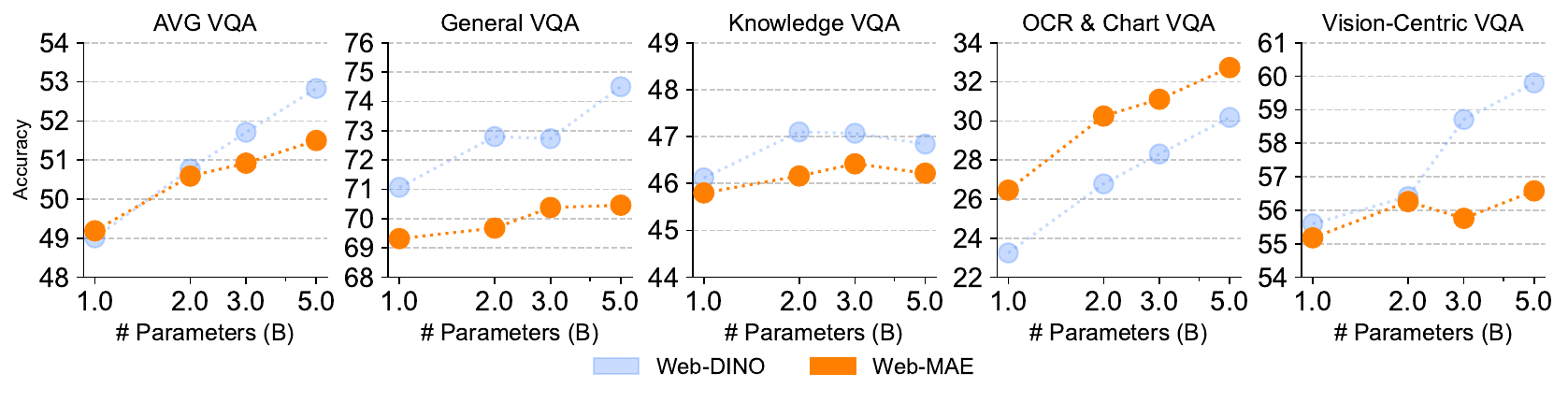}
\caption{
\textbf{\ourmae{} trained on \ourdata{}.} \ourmae{} also exhibits consistent scaling behavior as model size increases. Notably, \ourmae{} demonstrates better performance in OCR \& Chart tasks, achieving higher accuracy than \ourdino{} across all model sizes.}
    \label{fig:scale mae}
\end{figure*}

\researchquestion[generalize to other ssl]{Does the observed scaling behavior generalize to other visual SSL methods?}
\label{subsec:generalize to other ssl}

In previous sections, we derived our findings from DINOv2, a joint embedding visual SSL method. Here, we extend our analysis to a masked modelling based visual SSL method---Masked Autoencoder (MAE)~\citep{he2022masked}. We train MAE on \ourdata{} (denoted as \ourmae{}) using ViT models ranging from 1B to 5B parameters and compare the results with \ourdino{} models in \cref{fig:scale mae}.

\ourmae{} models exhibit similar scaling behavior to \ourdino{} models, with average VQA performance improving consistently as model size increases. Compared to joint embedding methods, \ourmae{} models learn features that are particularly well-suited for OCR \& Chart tasks but underperform in other domains. These results suggest that the ``scaling behavior'' observed in VQA tasks generalizes across different visual SSL methods. We also note that different visual SSL approaches learn distinct representations even when trained under the same conditions, as demonstrated by \ourmae{}'s OCR performance.

\researchquestion[small data]{Does visual SSL exhibit similar scaling behavior on smaller scale conventional data, such as ImageNet?}

\begin{figure*}[t]
    \centering
    \includegraphics[width=\linewidth]{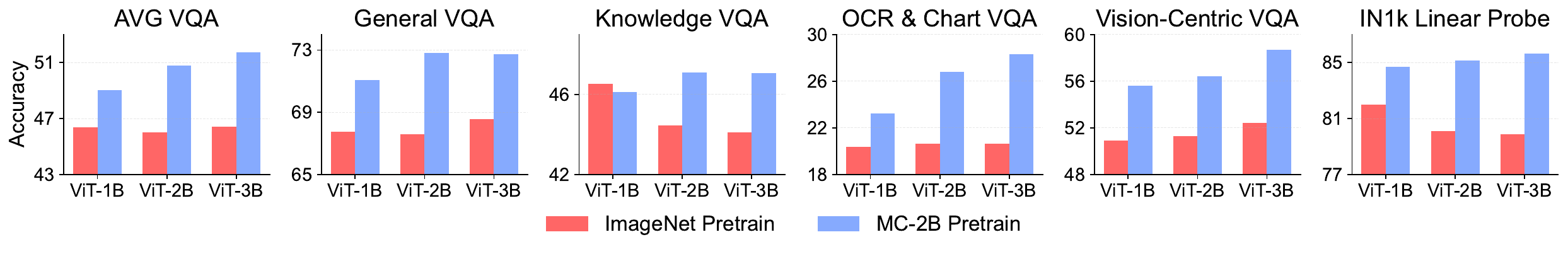}
    \caption{
    \textbf{Comparison of ImageNet-1k and \ourdata{} Pretraining.} Increasing the diversity and scale of pretraining data improves model performance on VQA accuracy and ImageNet linear probing. Unlike \ourdata{} pretraining, training on ImageNet does not exhibit a clear scaling trend.}
    \label{fig:smaller data}
    \vspace{-0.2cm}
\end{figure*}

We pretrain \ourdino{} 1B, 2B, and 3B models for 300 epochs on ImageNet-1k, a conventional pretraining dataset for SSL, following the recipe from \citep{oquab2023dinov2}. We compare these variants to those trained on \ourdata{}. We evaluate their downstream VQA performance and ImageNet-1k linear probing results. As shown in \cref{fig:smaller data}, models pretrained on ImageNet-1k exhibit consistently inferior performance across all the metrics. Moreover, unlike models trained on \ourdata{}, those trained on ImageNet-1k do not improve with increasing model sizes. This highlights the importance of training visual SSL on more diverse and larger datasets. This echoes recent findings that increasing dataset sizes and diversity drive LLM scaling~\citep{kaplan2020scaling, hoffmann2022training, chowdhery2022palm}, and also that pretraining data distribution is critical to downstream performance \citep{liu2024decade}.

\researchquestion[classic vision]{How do scaled models perform on classic vision tasks?}
\begin{figure*}[t]
    \centering
    \includegraphics[width=\linewidth]{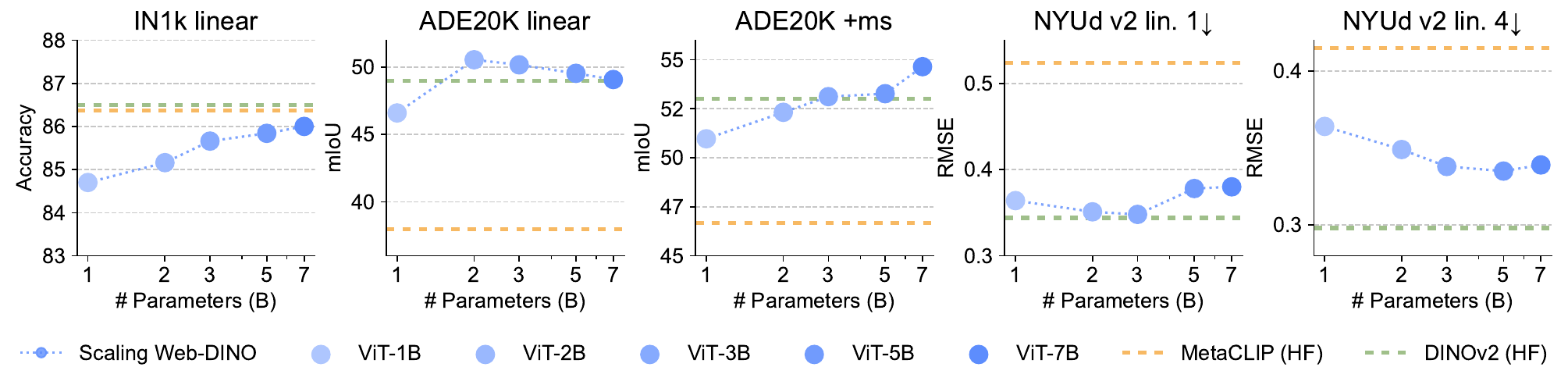}
    \caption{
    \textbf{Performance of \ourdino{} models on classic vision tasks.} All models achieve strong performance across ImageNet-1k classification, ADE20K segmentation, and NYU Depth estimation, and all tasks experience moderate improvements from increasing model size from 1B to 7B parameters. \ourdino{} outperforms MetaCLIP (HF) and is competitive with DINOv2 (HF). (HF) denotes the largest official Hugging Face released version.}
    \label{fig:classic benchmarks}
\end{figure*}

\definecolor{darkgreen}{rgb}{0.0, 0.5, 0.0}

We evaluate \ourdino{} models, ranging from 1B to 7B parameters, on classic vision benchmarks including linear probing on ImageNet-1k~\citep{deng2009imagenet}, semantic segmentation on ADE20K~\citep{zhou2019semantic}, and depth estimation on NYUv2~\citep{silberman2012indoor}. Following the evaluation protocol of DINOv2~\citep{oquab2023dinov2}, we freeze the vision encoder; see \cref{appdendix:implementation details} for details. As shown in \cref{fig:classic benchmarks}, \ourdino{}'s performance improves modestly with increasing model size. \ourdino{} achieves strong performance across all benchmarks, outperforming MetaCLIP by a significant margin and remaining competitive with off-shelf DINOv2, even outperforming it on ADE20K +ms. Note that the comparison with off-shelf DINOv2 is not exactly apples-to-apples, as we do not use high-resolution adaptation~\citep{oquab2023dinov2}, in order to maintain the same input resolution as CLIP. Additionally, the DINOv2 training data has a higher correlation with these classic vision benchmarks, detailed further in \cref{appendix:pretrain_dataset}. These differences suggest that there remains considerable room for further improvement in our model's classic vision performance.

However, we observe that the scaling behavior in classic vision tasks is less pronounced compared to VQA. This finding, along with insights from previous work~\citep{tong2024cambrian, fini2024multimodal, naeem2024silc}, reinforces the value of VQA as a comprehensive vision model evaluation framework. While classic benchmarks remain important, VQA provides a complementary view into model performance via offering a diverse set of tasks that are grounded in real-world perceptual challenges.

\begin{figure}[t]
    \centering
    \includegraphics[width=\linewidth]{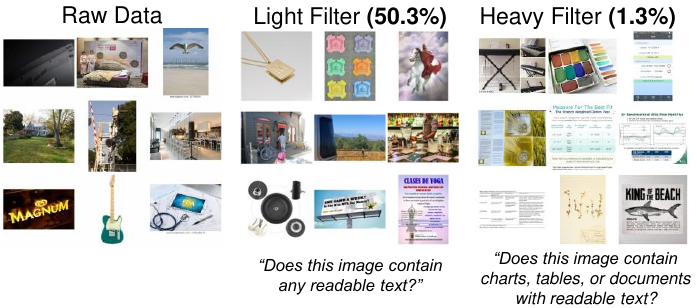}
    \caption{
    \textbf{Examples of filtered \ourdata{} images.} The Light filter (\textit{Middle}) identifies images containing text, retaining 50.3\% of the images. The Heavy filter (\textit{Right}) identifies images explicitly containing charts and documents, retaining only 1.3\% of \ourdata{}.}
    \label{fig:filter text}
    \vspace{-0.2cm}
\end{figure}

\newcommand{\plusvalue}[1]{\hspace{0.3em}\textcolor{darkgreen}{{\fontsize{7pt}{8pt}\selectfont\textbf{(+#1)}}}}
\newcommand{\minusvalue}[1]{\hspace{0.3em}\textcolor{red}{{\fontsize{7pt}{8pt}\selectfont\textbf{(-#1)}}}}

\begin{table*}[t!]
    \centering
    \fontsize{7.5pt}{8.0pt}\selectfont
    \setlength\tabcolsep{3.0pt} 
    \renewcommand{\arraystretch}{1.1} 
    \scalebox{1.00}{
    \begin{tabular}{rr|lllll |llll  }
 & &
     \multicolumn{5}{c|}{VQA Evaluator} &
     \multicolumn{4}{c}{Breakdown of OCR \& Chart Tasks} \\
      Method & \shortstack{\% of \\ \ourdata{}} & 
\rotatebox{0}{AVG} &
\rotatebox{0}{General} &
\rotatebox{0}{Knowledge} &
\rotatebox{0}{\shortstack{Vision\\Centric}} &
\rotatebox{0}{\shortstack{OCR\\Chart}} &
\rotatebox{0}{ChartQA}&
\rotatebox{0}{OCRBench} &
\rotatebox{0}{TextVQA} &
\rotatebox{0}{DocVQA} \\
      \hline

        \color{gray!50} 
CLIP 2B& 100\% & \color{gray!50}53.0 & \color{gray!50}72.2 & \color{gray!50}48.8 & \color{gray!50}55.0 & \color{gray!50}36.1 & \color{gray!50}32.8 & \color{gray!50}32.9& \color{gray!50}52.6& \color{gray!50}26.0\\ 
      \ourdino{} 2B & 100\% & 50.8 & 72.8 & 47.1 & 56.4 & 26.8 & 23.3& 15.6& 49.2& 19.0 \\
\ourdino{} 2B & 50.3\% & 53.4\plusvalue{2.6} & 73.0\plusvalue{0.2} & 51.7\plusvalue{4.6} & 55.6\minusvalue{0.8} & 33.2\plusvalue{6.4} & 31.4\plusvalue{8.1} & 27.3\plusvalue{11.7} & 51.3\plusvalue{2.1} & 23.0\plusvalue{4.0} \\
\ourdino{} 2B & 1.3\% & 53.7\plusvalue{2.9} & 70.7\minusvalue{2.1} & 47.3\plusvalue{0.2} & 56.2\minusvalue{0.2} & 40.4\plusvalue{13.6} & 47.5\plusvalue{24.2} & 29.4\plusvalue{13.8} & 52.8\plusvalue{3.6} & 32.0\plusvalue{13.0} \\

    \end{tabular}
}

\caption{
\textbf{Impact of data filtering on SSL model performance.} We compare Web-DINO ViT-2B models trained on \ourdata{} with different levels of text filtering (full, 50.3\%, and 1.3\%) against CLIP ViT-2B trained on full \ourdata{}. OCR \& Chart performance improves with progressively aggressive filtering, with the 1.3\% filter achieving the best results. Despite receiving zero language supervision, SSL models can surpass CLIP in text-centric tasks while maintaining strong overall performance.}

\label{tab:filter_text}
\end{table*}

\researchquestion[probe into text]{Why does web-scale data improve OCR \& Chart performance?}

In \cref{sec:exp}, we observed that increasing model size and examples seen leads to unprecedented improvements in OCR \& Chart performance for visual SSL models. This is surprising since current off-the-shelf visual SSL methods are notably poor at OCR \& Chart understanding compared to language-supervised models~\citep{tong2024cambrian, shi2024eagle}. 

One possible explanation is that web-scale image datasets already contain a degree of textual information. Unlike object-centric datasets such as ImageNet, images from the web often contain text (e.g. labels, signs, diagrams, etc.). Larger capacity and more data might aid visual SSL models to extract and leverage this textual information.

To test this hypothesis, we apply an off-the-shelf MLLM---SmolVLM2~\citep{allal2025smollm2}---to identify images containing text.
See \cref{fig:filter text} for qualitative examples and \cref{appdendix:implementation details} for details. This results in two curated datasets:
(i) Light filter: retains 50.3\% of \ourdino{} and contains images with any textual content. (ii) Heavy filter: retains 1.3\% of \ourdata{} and contains images with charts, tables, or documents.

We train \ourdino{} ViT-2B models on these filtered datasets, with each experiment using 2 billion seen examples (meaning filtered datasets undergo multiple epochs). As shown in \cref{tab:filter_text}, the model trained on lightly filtered data outperforms the full data variant by +6.4\% on OCR \& Chart, while maintaining strong performance in other categories. The model trained on heavily filtered data performs better and outperforms even the language-supervised CLIP ViT-2B trained on full data by +4.3\% on OCR \& Chart. Likewise, heavy filtering also improves Average VQA performance, outperforming the full data \ourdino{} ViT-2B by +2.6\% and even the full data CLIP ViT-2B by +0.7\%. This means that is it possible for visual SSL models to outperform CLIP models of the same size, with only a fraction of the total data (in this case 1.3\% of \ourdata{}).

The improvement in OCR \& Chart from training on heavily filtered data is particularly pronounced for ChartQA (+24.2\%), OCRBench (+13.8\%), and DocVQA (+13.0\%), while performance remains competitive in all other categories. These results demonstrate that self-supervised visual models, when trained on images containing more text in them, can develop high-quality text understanding capabilities without language supervision. It suggests that data composition---rather than purely scale or language supervision---is crucial for developing strong OCR \& Chart understanding abilities.

Although it is not surprising that skewing the data in favor of OCR \& Chart would improve OCR \& Chart capabilities, it is surprising that simple data filtering can outperform language supervision on the full data. This simple proof of concept suggests that similar techniques may be used to help visual SSL bridge future gaps in other capabilities.

\researchquestion[properties]{Why can SSL learn strong visual representations for multimodal modeling, without language supervision?}

\begin{figure}[t]
    \centering
    \includegraphics[width=1.02\linewidth]{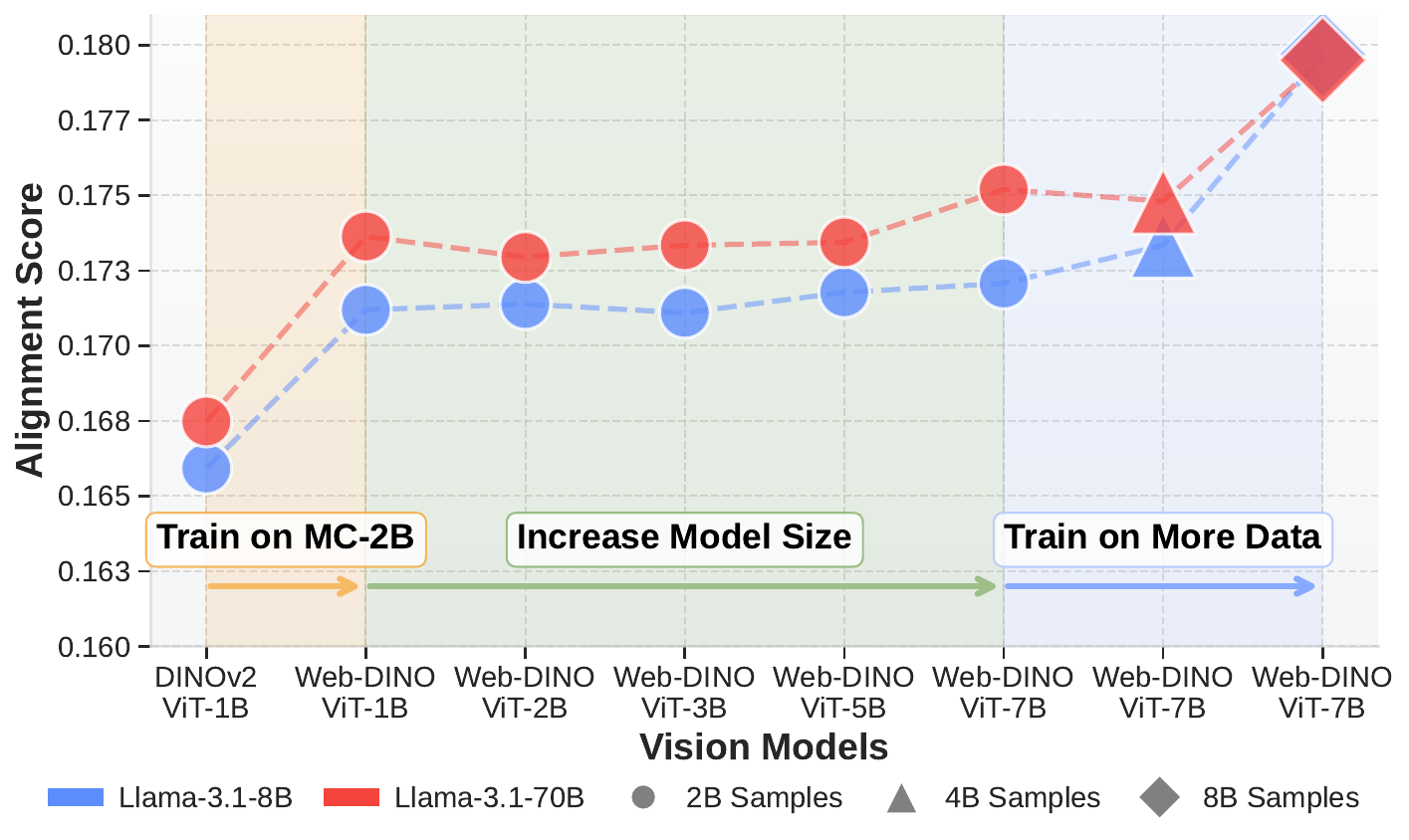}
    \caption{
    \textbf{Alignment score between \ourdino{} and LLMs.} 
    Moving from DINOv2 to \ourdino{} improves the alignment between the image and the corresponding text representations obtained by LLMs. Increasing model size from 1B to 7B parameters shows gradual improvement, while training on larger data quantities (4B/8B samples) yields the most significant alignment gains.}
    \label{fig:platonic benchmarks}
    \vspace{-0.2cm}
\end{figure}

Thus far, we have seen that visual SSL models can not only become competitive with CLIP models, but also that they can excel at tasks previously thought to require language. This raises an important question: why do vision-only models learn features that work well for multimodal models, even in the absence of language supervision? 

We hypothesize that SSL models learn features increasingly aligned with language as model size and examples seen increases. Following \citet{huh2024platonic}, we evaluate intrinsic representational alignment by computing a matching metric between the vision encoder and language model, using image-text pairs from the Wikipedia Captions dataset \citep{srinivasan2021wit}. We use off-the-shelf DINOv2~\citep{oquab2023dinov2} and \ourdino{} as vision encoders, and off-the-shelf Llama-3.1 8B and 70B~\citep{touvron2023llama} as the language models, \textit{without} any visual instruction tuning nor alignment procedure.

As shown in \cref{fig:platonic benchmarks}, we observe three key trends: (1) training on more diverse data (\ourdata{}) improves alignment with LLMs (DINOv2 ViT-1B → \ourdino{} ViT-1B); (2) increasing the vision model size leads to slightly higher alignment (\ourdino{} ViT-1B → ViT-7B); and (3) seeing more training samples further enhances alignment (\ourdino{} ViT-7B trained on 2B samples → 8B samples).

These findings suggest that as model size and, in particular, training samples scale, vision models naturally develop text-sensitive features and achieve strong alignment with LLMs and multimodal tasks, without explicit language supervision. 

\definecolor{blue3}{HTML}{5D8DFD}
\definecolor{green3}{HTML}{88B06D}
\definecolor{orange3}{HTML}{F5A83D}
\definecolor{red3}{HTML}{F5433D}
\definecolor{blue1}{HTML}{AEC6FE}
\definecolor{green1}{HTML}{B3CDA2}
\definecolor{orange1}{HTML}{F9CB8A}
\definecolor{red1}{HTML}{FBA09D}
\definecolor{lightgray}{gray}{1.0}
\definecolor{cambriangray}{gray}{0.9}

\begin{table*}[t]
    \centering
    \fontsize{7.2pt}{8.0pt}\selectfont
    \setlength\tabcolsep{5.0pt} 
    \renewcommand{\arraystretch}{1.5} 
    \scalebox{1.00}{
    \begin{tabular}{rccl|ccccc |ccccc  }
     \multicolumn{4}{c|}{Model} &
     \multicolumn{5}{c|}{MLLM Evaluator} &
     \multicolumn{5}{c}{Classic Vision Tasks} \\
      Method & \rotatebox{0}{\shortstack{Pretrain \\Data}}&
      \rotatebox{0}{\shortstack{Pretrain \\Samples \\Seen}} &
      \rotatebox{0}{Res} &
      \rotatebox{90}{AVG} &
      \rotatebox{90}{General} &
      \rotatebox{90}{Knowledge} &
      \rotatebox{90}{OCR \& Chart} &
      \rotatebox{90}{Vision-Centric} &
      \rotatebox{90}{IN1k lin.} &
      \rotatebox{90}{ADE20K lin.} &
      \rotatebox{90}{ADE20K ms.} &
        \rotatebox{90}{NYUd lin. 1 ($\downarrow$)} & 
        \rotatebox{90}{NYUd lin. 4 ($\downarrow$)}
  \\
      \hline
    \rowcolor{gray!10}   
    \multicolumn{4}{l|}{\textbf{Language-Supervised Models}} &  &  &  &  &  &   &  &  & & \\
    
    \multirow{2}{*}{SigLIP ViT-SO400M} 
      & \multirow{2}{*}{WebLI} & \multirow{2}{*}{45.0B} & 224 
      & 55.4 & 74.4 & 48.7 & 39.5 & 58.9 
      & 86.5 & 36.5 & 38.0 & 0.607 & 0.525\\ 
    & & & 384
      & 60.0 & 76.3 & 50.4 & 53.5 & 59.7 
      & 87.3 & 39.5 & 47.2 & 0.582 & 0.438\\ 
    \arrayrulecolor{gray!40}\cline{1-14}\arrayrulecolor{black}
    
    \multirow{2}{*}{SigLIP2 ViT-SO400M} 
      & \multirow{2}{*}{WebLI} & \multirow{2}{*}{45.0B} & 224 
      & 56.3 & 74.4 & 50.7 & 42.1 & 58.1 
      &   87.5   & 41.1     &   44.2   & 0.562 & 0.539\\ 
    & & & 384
      & 62.0 & 76.6 & 51.9 & 58.4 & 61.0

      & 88.1 & 43.5 & 50.2 & 0.524 & 0.469 \\ 
    \arrayrulecolor{gray!40}\cline{1-14}\arrayrulecolor{black}
    
    MetaCLIP ViT-G 
      & MetaCLIP & 12.8B & 224
      & 54.8 & 75.5 & 48.2 & 37.3 & 58.4
      & 86.4 & 38.0 & 46.7 & 0.524 & 0.415\\
      
    \rowcolor{gray!10}   
    \multicolumn{4}{l|}{\textbf{Visual Self-Supervised Models}} &  &  &  &  &  &   &  &  & &  \\
    
    MAE ViT-H 
      & ImageNet-1k & 2.0B & 224
      & 45.2 & 64.6 & 43.9 & 20.6 & 51.7
      & 76.6 & 33.3 & 30.7 & 0.517 & 0.483\\
    \arrayrulecolor{gray!40}\cline{1-14}\arrayrulecolor{black}
    
    I-JEPA ViT-H 
      & ImageNet-22k & 0.9B & 224
      & 44.7 & 65.4 & 43.9 & 21.2 & 48.4
      & 68.8 & 31.6 & 34.6 & 0.548 & 0.520\\
    \arrayrulecolor{gray!40}\cline{1-14}\arrayrulecolor{black}
    

        DINOv2 ViT-g
      & LVD-142M & 1.9B & 518
      & 47.9 & 70.2 & 45.0 & 21.2 & 55.3
      &
      86.0 & 49.0 & 53.0 & 0.344 & 0.298  \\
      
    \arrayrulecolor{gray!50}\cline{1-14}\arrayrulecolor{black}
    
    \rowcolor{green!5}
    &  & & 224 
      & 55.2 & 74.5 & 48.0 & 39.4 & 59.1
      & 86.5 & 42.1 & 52.6 & 0.491 & 0.376\\
    \rowcolor{green!5}
     &  &  & 378
      & 57.4 & 73.9 & 47.7 & 50.4 & 57.7
      & 86.3 & 42.3 & 53.1 & 0.498 & 0.366 \\
    \rowcolor{green!5}
     \multirow{-3}{*}{Web-DINO ViT-7B} & \multirow{-3}{*}{\ourdata{}}  & \multirow{-3}{*}{8.0B}  & 518
      & 59.9 & 75.5 & 48.2 & 55.1 & 60.8
      & 86.4 & 42.6 & 52.8 & 0.490 & 0.362 \\

    \end{tabular}
    }
 \caption{\textbf{Comparison with other vision models.} 
 Web-DINO ViT-7B achieves competitive performance with CLIP models on VQA without language supervision and surpasses them on traditional vision tasks. 
 Compared to other self-supervised models like DINOv2, Web-DINO significantly narrows the performance gap with CLIP on VQA tasks, 
 particularly excelling in OCR \& Chart understanding. 
 These results demonstrate that SSL can effectively produce strong visual representations for both multimodal and classic vision tasks.}
\label{tab:final_table}
\end{table*}

\section{The \ourssl{} Model Family}
\label{sec:final model}

Next, we analyze the overall best performing vision encoders using both VQA and classic vision benchmarks. In \Cref{tab:final_table}, we show the best results of our vision encoders against recent off-the-shelf vision encoders, in terms of VQA and classic vision tasks.

For VQA, all vision encoders---including off-the-shelf models---are evaluated using the same visual instruction tuning setup detailed in \cref{subsec:MLLM as evaluation}, and mainly 224$\times$224 input resolution for the purpose of fair comparison. Because the goal is not to produce a state-of-the-art MLLM, we did not employ techniques such as unfreezing the vision encoder, resolution tiling~\citep{liu2024llavanext}, and spatial visual aggregator~\citep{tong2024cambrian}.

For classic vision, we follow the evaluation procedure from \citet{oquab2023dinov2} and evaluate linear probe performance on ImageNet-1k~\citep{deng2009imagenet}, ADE20K~\citep{zhou2019semantic}, and NYU Depth v2~\citep{silberman2012indoor}. The input resolution differs between classic vision tasks, but each model tested uses the same exact settings from \citet{oquab2023dinov2}. We emphasize that the primary motivation is still to provide controlled insights.

\noindent\textbf{Performance at 224px.}  \ourdino{} can outperform off-the-shelf MetaCLIP in both VQA and classic vision tasks. \ourdino{} is even able to match the performance of SigLIP and SigLIP2 on VQA despite seeing 5$\times$ less data and receiving no language supervision. In general, \ourdino{} outperforms all off-shelf language-supervised CLIP models at traditional vision benchmarks. Although our best \ourdino{} model is 7B parameters, the results from \cref{subsec:scale vit} and \cref{subsec:scale data} suggest that CLIP models saturate beyond moderate model and data sizes, while visual SSL improves progressively with increasing model and data size. \ourdino{} also outperforms off-the-shelf visual SSL methods, including DINOv2~\citep{oquab2023dinov2}, in all VQA categories. \ourdino{} is also competitive in traditional vision benchmarks.

\noindent\textbf{Performance beyond 224px.} 
Next, we discuss the performance of higher resolution models. Following \citet{oquab2023dinov2}, we additionally fine-tune \ourdino{} for 20k steps. We do this for resolutions of 378 and 518, to compare against the higher-resolution off-shelf versions of SigLIP as well as DINO. See \cref{appendix:high res} for training details. From 224 to 378 to 518 resolution, \ourdino{} improves steadily at average VQA, with notable gains in OCR \& Chart performance. Classic vision performance improves modestly with higher resolution. At 384 resolution, \ourdino{} trails behind SigLIP. At 518 resolution, \ourdino{} is largely able to bridge the gap. The results suggest that \ourdino{} may benefit from further increasing high-resolution adaptation.

\section{Related Work}
\label{sec:related_work}
\paragraph{Visual self-supervised learning methods.}
Early visual SSL methods explored various pretext tasks for pretraining~\citep{wang2015unsupervised, doersch2015unsupervised, noroozi2016unsupervised, zhang2016colorful, gidaris2018unsupervised, balestriero2023cookbook}. More recently, research has converged on two primary approaches: joint embedding methods and masked image modeling. Joint embedding methods learn invariant features by aligning representations of different augmented views~\citep{he2019momentum, misra2019self, chen2020simple, grill2020bootstrap, chen2020improved, chen2021exploring, chen2021empirical, caron2021emerging,lecun2022path, chen2022bag, garrido2022duality}, while masked modeling~\citep{zhou2021ibot, he2022masked, wei2022masked, fan2023motion, assran2023self,woo2023convnext,barstochastic,bai2024sequential,carreira2024scaling} learns by predicting masked visual inputs.

Our work complements SSL research focused on pretraining algorithms, by taking off-the-shelf training code and training visual SSL at scale with a controlled experimental setup. In Question~\ref{rq:generalize to other ssl}, we show that the observed scaling behavior generalizes across both joint embedding and masked modeling SSL methods, and is likely not a method-specific phenomena.

\paragraph{Data used to train vision models.}
Both supervised \citep{he2016resnet, xie2016resnext, dosovitskiy2020image, liu2022convnet} and SSL vision models have traditionally relied on standard datasets such as MNIST \citep{lecun1998mnist}, CIFAR-10 \citep{krizhevsky2009learning}, and ImageNet \citep{deng2009imagenet, ridnik2021imagenet}. More recently, self-supervised methods have scaled to larger unlabeled datasets, such as YFCC \citep{thomee2016yfcc100m}, LVD-142M \citep{oquab2023dinov2}, and IG-3B \citep{singh2023effectiveness}; however, these methods still exhibit a significant performance gap compared to language-supervised models on VQA.

In contrast, language-supervised models~\citep{radford2021learning, zhai2023sigmoid,sun2023eva, sun2024eva, xu2023demystifying, tang2025tulip} leverage significantly larger image-text datasets, from WIT-400M \citep{radford2021learning} to billion-scale web data~\citep{schuhmann2022laion, fang2023data, xu2023demystifying, gadre2024datacomp}, with some using up to 100B image-text pairs \citep{wang2025scaling}. Studies suggest that pretraining data distribution is more critical for downstream performance than specific training methodologies \citep{fang2022data, liu2024decade}.

Our work bridges these paradigms by pretraining SSL models on web-scale data. Through controlled experiments (\cref{sec:exp} and \ref{sec:analysis}), we show that (1) visual SSL models are sensitive to the training distribution, (2) increasing data diversity and quantity significantly improves performance on a diverse range of VQA tasks, and (3) training on a higher concentration of images containing text is highly effective for improving OCR \& Chart understanding.

\paragraph{Evaluating vision models.}
Classic works have primarily used image classification~\citep{lecun1998mnist, krizhevsky2009learning, deng2009imagenet, bossard2014food, hendrycks2019natural, hendrycks2020many} to evaluate learned representations. More recent SSL research has expanded evaluation to include image segmentation~\citep{everingham2010pascal, cordts2016cityscapes, he2017mask,zhou2019semantic}, depth estimation~\citep{silberman2012indoor, geiger2013vision, song2015sun}, and video classification~\citep{soomro2012ucf101, goyal2017something, dehghan2021arkitscenes}. Language-supervised models~\citep{radford2021learning, zhai2023sigmoid}, due to their two-tower encoder structure, commonly use zero-shot image classification to assess the quality of learned image and text features.

Our work follows recent proposals~\citep{naeem2024silc, fini2024multimodal, tong2024cambrian} to evaluate vision encoders on a broader range of VQA tasks~\citep{goyal2017making, yue2023mmmu, liu2023mmbench, fu2023mme, tao2024what,yue2024mmmu,grok} using MLLMs. These VQA tasks complement traditional vision benchmarks by assessing visual features on a more diverse range of real-world perceptual challenges. As shown in \cref{sec:exp} and \cref{sec:analysis}, we find that visual SSL trained on web-scale data learns representations that continue to improve on VQA benchmarks, and---to a lesser degree---also on traditional vision benchmarks.

\section{Limitations}
\label{sec:limitations}
In this work, we focus on training visual SSL models without using language. The main limitation of vision-only models, compared to language-supervised models, is that they do not support zero-shot image classification out of the box. However, by integrating visual SSL models into MLLM frameworks through instruction tuning, we show they can achieve impressive downstream performance across classification and other tasks. Another way to achieve zero-shot image classification is to use LiT-style adaptation \citep{zhai2022lit, jose2024dinov2}, but this is out-of-scope for our work as we do not use language supervision. To focus on comparing the vision encoder, we fixed the base LLM for visual instruction tuning to Llama-3 8B Instruct~\citep{llama3modelcard}. We hypothesize that the findings using other LLM backbones would be similar, however this is not in scope for our work. Additionally, while we demonstrate that visual SSL scales well on MetaCLIP data, we leave the exploration of even larger and/or uncurated datasets to future work.

\section{Discussion}
\label{sec:discussion}
We show that large-scale visual encoders that are trained with self-supervised language-free objectives can produce high quality visual features for multimodal models.
Our results echo the ``bitter lesson''~\citep{sutton2019bitter} and suggest that imposing less supervision---including language---remains a promising direction for advancing the field of computer vision. We hope our work will inspire further exploration of vision-only approaches, which will enable the construction of next generation vision models that excel at both traditional vision and modern multimodal capabilities.
 
\section{Acknowledgements}
\label{sec:acknowledgements}
We thank Ellis Brown, John Nguyen, Junlin Han, Shengyi Qian, Tyler Zhu, Yuexiang Zhai, Druv Pai, Shusheng Yang, Jihan Yang, Muzi Tao, Boyang Zheng, and Anjali Gupta for reviewing this manuscript. We thank Hu Xu and the MetaCLIP paper authors for creating the MetaCLIP dataset. We thank Mido Assran, Mikael Henaff, Daniel Bolya, Hu Xu, Mark Ibrahim, Russ Howes, and Matthew Muckley for their insightful feedback. We thank Michaël Ramamonjisoa and Marc Szafraniec for their help with image segmentation and depth estimation evaluations. Lastly, we thank Ananya Saxena, Cody Olsen, Mack Ward, Maxwell Taylor, Kalyan Saladi, Dev Satpathy, Dinesh Kannappan, Xiaodong Ma, Jacob Kahn, Gabriel Synnaeve, and Shubho Sengupta for infrastructure support.

\clearpage

{
    \small
    \bibliographystyle{ieeenat_fullname}
    \bibliography{main}
}

\input{appendix.tex}

\end{document}

%% file: appendix.tex
\clearpage
\clearpage
\appendix
\section{Implementation Details} \label{appdendix:implementation details}
\paragraph{Training.}
For training \ourdino{}, \ourmae{}, and CLIP models, we closely follow the existing open-source codebases: the official DINOv2 and MAE repositories, and the MetaCLIP codebase which builds on top of the OpenCLIP codebase~\citep{cherti2023reproducible}.
We use Fully Sharded Data Parallel (FSDP)~\citep{zhao2023pytorch} for distributed training of larger models.

For \ourdino{} and CLIP pretraining, we follow the exact recipe and hyperparameters from the original paper for their largest model.
For MAE pretraining, we observe that training becomes more prone to divergence as model size increases. To mitigate this, we reduce the learning rate from 2.4e-3 to 1.6e-3 and extend the warmup period to 80K iterations.
Table~\ref{tab:training_config} provides a summary of the pretraining hyperparameters.

\begin{table}[h]
    \centering
    \small
    \vspace{-0.1cm}
    \setlength\tabcolsep{3pt} 
    \begin{tabular}{lcccc}
    \toprule
    Model & Batch Size  & Learning Rate & Warmup \\
    \midrule
    \ourdino{} & 3072  & 3.5e-4 & 100K  \\
    \ourmae{} & 4096  & 1.6e-3 & 80K   \\
    CLIP & 32768  & 4e-4 & 2K \\
    \bottomrule
    \end{tabular}
    
    \vspace{-0.1cm}
    \captionsetup{font=footnotesize}
    \caption{\footnotesize \textbf{Hyperparameters for \ourdino{}, \ourmae{} and CLIP.}}
    \label{tab:training_config}
    \vspace{-0.1cm}
\end{table}


\paragraph{VQA evaluation.}
For VQA evaluation, we follow \citet{tong2024cambrian, tong2024metamorph} and use Cambrian-Alignment data for MLP projector training and Cambrian-7M for MLP and LLM fine-tuning. We finetune on top of Llama-3 8B Instruct~\citep{llama3modelcard}. The vision encoder is frozen throughout finetuning. We excluded LAION~\citep{schuhmann2022laion} images from the Cambrian data to comply with safety standards. We first encode the images at the model's original input resolution using the pretrained vision encoder. Next, we extract features from the final encoder layer. Following prior approaches~\citep{tong2024cambrian, tong2024metamorph}, we then resize the resulting token sequence to a fixed length of 576 tokens through bilinear interpolation. This ensures consistency across evaluations despite variations in input image resolutions. We report configurations in \cref{tab:cambrian training}.

\begin{table*}[h]
\centering
    \small  
    \setlength\tabcolsep{2pt} 
    \begin{adjustbox}{max width=\textwidth}
    \begin{tabular}{l|cc|ccc|ccc}
    
 \multicolumn{1}{c|}{Backbone} & \multicolumn{2}{c|}{Data} & \multicolumn{3}{c|}{Adapter} &  \multicolumn{3}{c}{Instruction Tuning}\\
\multicolumn{1}{c|}{LLM} & Adapter & Instruction Tuning &     LR & WD & BS & LR & WD & BS \\
      \hline
 Llama-3 8B Instruct& Cambrian Adapter Data &  Cambrian-7M & 1.00e-5 & 0.0 & 512 & 4.00e-5 & 0 & 512 \\
    \end{tabular}
    \end{adjustbox}
         \vspace{-1em}
\caption{\small \textbf{Hyperparameters for all VQA experiments.} We exclude LAION~\citep{schuhmann2022laion} from Cambrian data. 
}
\label{tab:cambrian training}
\end{table*}

\paragraph{Classic vision evaluation.}

We follow the evaluation procedure in DINOv2~\citep{oquab2023dinov2} for all classic vision evaluation: linear probe on ImageNet1k~\citep{deng2009imagenet}, ADE20K~\citep{zhou2019semantic}, and NYU Depth v2~\citep{silberman2012indoor}. For ImageNet-1k, we evaluate models with their pretrained image resolution; For ADE20K and NYU Depth v2, we use the settings from \citet{oquab2023dinov2}. For ADE20K, we follow DINOv2 and report the \textbf{linear} and \textbf{+ms} setting. For NYU Depth v2, we report \textbf{lin. 1} and \textbf{lin. 4}. See the original paper for additional details.

\paragraph{Model architectures.}
In \cref{tab:vit config}, we defined the ViT architectures used in our study. To recap, we first borrowed the ViT-g architecture from~\citet{oquab2023dinov2} and named it ViT-1B for consistent notation. We then define 2B, 3B, 5B, and 7B architectures inspired by language model scaling. Specifically, the 2 - 7B architectures are wider than the 1B variant, inspired by language model recipes. Our 7B architecture is almost identical to the Llama-2 7B design, except for the patch embedding layer which is unique to ViTs.

\paragraph{Text filtering.}
In Question~\ref{rq:probe into text}, we introduced the ``Light'' and ``Heavy'' filters which retain 50.3\% and 1.3\% of \ourdata{} respectively. Specifically, we use a small MLLM, SmolVLM2~\citep{allal2025smollm2}, to identify images containing text, using prompts such as \textit{``Does this image contain any readable text?''}. The intention is not to achieve perfect filtering, but rather to skew the data distribution in the general desired direction. See \cref{fig:filter text} for a visualization of the filtering process and some examples. This results in two curated datasets:

(i) Light filter: Retains 50.3\% of the original data, primarily consisting of images with some textual content. Prompt used: \textit{``Does this image contain any readable text? Answer only yes or no.''}

(ii) Heavy filter: Retains only 1.3\% of the data, focusing mainly on charts and documents. Prompt used: \textit{``Please think carefully before answering. Does this image contain charts, tables, or documents with readable text? Answer only yes or no.''}

\section{Full Results} \label{appendix:full results}

We include full results of all experiments presented in \cref{sec:exp} and \cref{sec:analysis}. 

\subsection{\ourdino{}}

\paragraph{Scaling up model sizes.} We show quantitative results of scaling up the model under VQA evaluation in \cref{tab:VQA PIN-DINO MC-2B Data-2B} and classic vision evaluation in \cref{tab:Vision PIN-DINO MC-2B Data-2B}. These are the numerical results for \cref{subsec:scale vit}.

\begin{table*}[ht]
\centering
    \small  
    \setlength\tabcolsep{2pt} 
    \begin{adjustbox}{max width=\textwidth}
    \begin{tabular}{l|r|rrrr|rrrr|rrrr|rrrr}
     \multicolumn{1}{c|}{Vision Backbone} &  \multicolumn{1}{c|}{} & \multicolumn{4}{c|}{General} & \multicolumn{4}{c|}{Knowledge} & \multicolumn{4}{c|}{OCR \& Chart} & \multicolumn{4}{c}{Vision-Centric}  \\
      Model  & \multicolumn{1}{c|}{\rotatebox{90}{Average}} & \multicolumn{1}{c}{\rotatebox{90}{MME$^\text{P}$}} & \multicolumn{1}{c}{\rotatebox{90}{MMB}} & \multicolumn{1}{c}{\rotatebox{90}{SEED$^\text{I}$}} & \multicolumn{1}{c|}{\rotatebox{90}{GQA}} & \multicolumn{1}{c}{\rotatebox{90}{SQA$^\text{I}$}} & \multicolumn{1}{c}{\rotatebox{90}{MMMU$^\text{V}$}} & \multicolumn{1}{c}{\rotatebox{90}{MathVista$^\text{M}$}} & \multicolumn{1}{c|}{\rotatebox{90}{AI2D}} & \multicolumn{1}{c}{\rotatebox{90}{ChartQA}} & \multicolumn{1}{c}{\rotatebox{90}{OCRBench}} & \multicolumn{1}{c}{\rotatebox{90}{TextVQA}} & \multicolumn{1}{c|}{\rotatebox{90}{DocVQA}} & \multicolumn{1}{c}{\rotatebox{90}{MMVP}} & \multicolumn{1}{c}{\rotatebox{90}{RealWorldQA}} & \multicolumn{1}{c}{\rotatebox{90}{CV-Bench$^\text{2D}$}} & \multicolumn{1}{c}{\rotatebox{90}{CV-Bench$^\text{3D}$}} \\
\hline 
\ourdino{} ViT-1B & 49.01 & 1731.52 & 65.37 & 69.92 & 62.40 & 72.58 & 35.33 & 12.30 & 64.28 & 19.20 & 9.40 & 47.41 & 17.00 & 37.33 & 57.12 & 64.80 & 63.16 \\
\ourdino{} ViT-2B & 50.77 & 1760.80 & 68.98 & 71.29 & 62.89 & 73.67 & 31.77 & 15.90 & 67.06 & 23.30 & 15.60 & 49.20 & 19.00 & 38.00 & 57.38 & 65.85 & 64.41\\
\ourdino{} ViT-3B & 51.71 & 1757.27 & 68.04 & 71.84 & 63.19 & 73.57 & 33.00 & 14.40 & 67.32 & 25.68 & 17.10 & 50.45 & 20.00 & 42.66 & 56.86 & 69.49 & 65.83 \\
\ourdino{} ViT-5B & 52.83 & 1840.81 & 70.01 & 72.39 & 63.56 & 75.06 & 32.11 & 12.40 & 67.77 & 26.96 & 22.10 & 50.64 & 21.00 & 44.66 & 57.64 & 67.75 & 69.16 \\
\ourdino{} ViT-7B & 53.87 & 1823.76 & 68.98 & 73.02 & 64.22 & 74.61 & 35.11 & 14.00 & 69.43 & 28.80 & 23.59 & 51.10 & 22.00 & 48.00 & 59.34 & 69.96 & 68.58
    \end{tabular}
\end{adjustbox}
\caption{\textbf{VQA Evaluation: \ourdino{} trained on \ourdata{} with 2 billion images seen.}}
\label{tab:VQA PIN-DINO MC-2B Data-2B}
\end{table*}

\begin{table*}[ht]
\centering
    \small  
    \setlength\tabcolsep{6pt} 
    \begin{adjustbox}{max width=\textwidth}
    \begin{tabular}{l|ccccc}
     \multicolumn{1}{c|}{Vision Backbone} & \multicolumn{1}{c}{\rotatebox{0}{IN1k lin.}} & \multicolumn{1}{c}{\rotatebox{0}{ADE20K lin.}} & \multicolumn{1}{c}{\rotatebox{0}{ADE20K +ms.}} & \multicolumn{1}{c}{\rotatebox{0}{NYUd lin. 1 (↓)} } & \multicolumn{1}{c}{\rotatebox{0}{NYUd lin. 4 (↓)}} \\
\hline 
\ourdino{} ViT-1B & 84.70 & 46.60 & 50.97 & 0.364 & 0.345 \\
\ourdino{} ViT-2B & 85.16 & 50.55 & 52.32 & 0.351 & 0.335 \\
\ourdino{} ViT-3B & 85.66 & 50.17 & 53.12 & 0.348 & 0.328 \\
\ourdino{} ViT-5B & 85.84 & 49.54 & 53.27 & 0.378 & 0.335 \\
\ourdino{} ViT-7B & 86.00 & 49.08 & 54.65 & 0.380 & 0.339 \\
    \end{tabular}
\end{adjustbox}
\caption{\textbf{Classic Vision Evaluation: \ourdino{} trained on \ourdata{} with 2 billion images seen.}}
\label{tab:Vision PIN-DINO MC-2B Data-2B}
\end{table*}

\paragraph{Scaling up data sizes.} We show quantitative results of scaling up the number of data seen with \ourdino{} ViT-7B on VQA evaluation in \cref{tab:VQA PIN-DINO MC-2B Data-scale} and classic vision evaluation in \cref{tab:Vision PIN-DINO MC-2B Data-scale}. These are the numerical results for \cref{subsec:scale data}.

\begin{table*}[ht]
\centering
    \small  
    \setlength\tabcolsep{2pt} 
    \begin{adjustbox}{max width=\textwidth}
    \begin{tabular}{l|r|rrrr|rrrr|rrrr|rrrr}
     \multicolumn{1}{c|}{Vision Backbone} &  \multicolumn{1}{c|}{} & \multicolumn{4}{c|}{General} & \multicolumn{4}{c|}{Knowledge} & \multicolumn{4}{c|}{OCR \& Chart} & \multicolumn{4}{c}{Vision-Centric}  \\
      Model  & \multicolumn{1}{c|}{\rotatebox{90}{Average}} & \multicolumn{1}{c}{\rotatebox{90}{MME$^\text{P}$}} & \multicolumn{1}{c}{\rotatebox{90}{MMB}} & \multicolumn{1}{c}{\rotatebox{90}{SEED$^\text{I}$}} & \multicolumn{1}{c|}{\rotatebox{90}{GQA}} & \multicolumn{1}{c}{\rotatebox{90}{SQA$^\text{I}$}} & \multicolumn{1}{c}{\rotatebox{90}{MMMU$^\text{V}$}} & \multicolumn{1}{c}{\rotatebox{90}{MathVista$^\text{M}$}} & \multicolumn{1}{c|}{\rotatebox{90}{AI2D}} & \multicolumn{1}{c}{\rotatebox{90}{ChartQA}} & \multicolumn{1}{c}{\rotatebox{90}{OCRBench}} & \multicolumn{1}{c}{\rotatebox{90}{TextVQA}} & \multicolumn{1}{c|}{\rotatebox{90}{DocVQA}} & \multicolumn{1}{c}{\rotatebox{90}{MMVP}} & \multicolumn{1}{c}{\rotatebox{90}{RealWorldQA}} & \multicolumn{1}{c}{\rotatebox{90}{CV-Bench$^\text{2D}$}} & \multicolumn{1}{c}{\rotatebox{90}{CV-Bench$^\text{3D}$}} \\
\hline 
\ourdino{} ViT-7B (1B Data) & 51.02 & 1785.97 & 68.12 & 72.54 & 63.60 & 73.87 & 32.88 & 12.70 & 66.58 & 23.60 & 15.20 & 49.04 & 19.00 & 43.33 & 57.12 & 68.35 & 61.08 \\
\ourdino{} ViT-7B (2B Data) & 53.87 & 1823.76 & 68.98 & 73.02 & 64.22 & 74.61 & 35.11 & 14.00 & 69.43 & 28.80 & 23.59 & 51.10 & 22.00 & 48.00 & 59.34 & 69.96 & 68.58 \\
\ourdino{} ViT-7B (4B Data) & 54.37 & 1827.12 & 71.39 & 72.61 & 63.53 & 72.73 & 34.00 & 18.90 & 67.09 & 35.12 & 30.00 & 53.19 & 24.00 & 45.33 & 55.94 & 69.68 & 65.00\\
\ourdino{} ViT-7B (8B Data) & 55.24 & 1811.05 & 71.30 & 72.14 & 64.04 & 72.43 & 35.66 & 15.20 & 68.52 & 35.52 & 36.40 & 56.53 & 29.00 & 46.00 & 57.90 & 70.53 & 62.08 \\

    \end{tabular}
\end{adjustbox}
\caption{\textbf{VQA Evaluation: \ourdino{} ViT-7B trained on \ourdata{} with increased number of images seen.}}
\label{tab:VQA PIN-DINO MC-2B Data-scale}
\end{table*}

\begin{table*}[ht]
\centering
    \small  
    \setlength\tabcolsep{6pt} 
    \begin{adjustbox}{max width=\textwidth}
    \begin{tabular}{l|ccccc}
     \multicolumn{1}{c|}{Vision Backbone} & \multicolumn{1}{c}{\rotatebox{0}{IN1k lin.}} & \multicolumn{1}{c}{\rotatebox{0}{ADE20K lin.}} & \multicolumn{1}{c}{\rotatebox{0}{ADE20K +ms.}} & \multicolumn{1}{c}{\rotatebox{0}{NYUd lin. 1 (↓)} } & \multicolumn{1}{c}{\rotatebox{0}{NYUd lin. 4 (↓)}} \\
\hline 
\ourdino{} ViT-7B (2B Data) & 86.00 & 49.08 & 54.65 & 0.380 & 0.339 \\
\ourdino{} ViT-7B (4B Data) & 86.33 & 47.41 & 54.66 & 0.416 & 0.363 \\
\ourdino{} ViT-7B (8B Data) & 86.52 & 42.14 & 52.55 & 0.491 & 0.376 \\
    \end{tabular}
\end{adjustbox}
\caption{\textbf{Classic Vision Evaluation: \ourdino{} ViT-7B trained on \ourdata{} with increased number of images seen.}}
\label{tab:Vision PIN-DINO MC-2B Data-scale}
\end{table*}

\paragraph{Scaling down training data.} We show VQA evaluation results from training \ourdino{} on less diverse data--ImageNet-1k, in \cref{tab:VQA PIN-DINO ImageNet Data-scale}. These are the full results for scaling down training data experiments in  Question~\ref{rq:small data}.

\begin{table*}[ht]
\centering
    \small  
    \setlength\tabcolsep{2pt} 
    \begin{adjustbox}{max width=\textwidth}
    \begin{tabular}{l|r|rrrr|rrrr|rrrr|rrrr}
     \multicolumn{1}{c|}{Vision Backbone} &  \multicolumn{1}{c|}{} & \multicolumn{4}{c|}{General} & \multicolumn{4}{c|}{Knowledge} & \multicolumn{4}{c|}{OCR \& Chart} & \multicolumn{4}{c}{Vision-Centric}  \\
      Model  & \multicolumn{1}{c|}{\rotatebox{90}{Average}} & \multicolumn{1}{c}{\rotatebox{90}{MME$^\text{P}$}} & \multicolumn{1}{c}{\rotatebox{90}{MMB}} & \multicolumn{1}{c}{\rotatebox{90}{SEED$^\text{I}$}} & \multicolumn{1}{c|}{\rotatebox{90}{GQA}} & \multicolumn{1}{c}{\rotatebox{90}{SQA$^\text{I}$}} & \multicolumn{1}{c}{\rotatebox{90}{MMMU$^\text{V}$}} & \multicolumn{1}{c}{\rotatebox{90}{MathVista$^\text{M}$}} & \multicolumn{1}{c|}{\rotatebox{90}{AI2D}} & \multicolumn{1}{c}{\rotatebox{90}{ChartQA}} & \multicolumn{1}{c}{\rotatebox{90}{OCRBench}} & \multicolumn{1}{c}{\rotatebox{90}{TextVQA}} & \multicolumn{1}{c|}{\rotatebox{90}{DocVQA}} & \multicolumn{1}{c}{\rotatebox{90}{MMVP}} & \multicolumn{1}{c}{\rotatebox{90}{RealWorldQA}} & \multicolumn{1}{c}{\rotatebox{90}{CV-Bench$^\text{2D}$}} & \multicolumn{1}{c}{\rotatebox{90}{CV-Bench$^\text{3D}$}} \\
\hline 
\ourdino{} ViT-1B & 46.39 & 1704.30 & 59.27 & 66.43 & 60.12 & 71.29 & 32.77 & 18.70 & 63.40 & 17.56 & 4.90 & 44.93 & 14.00 & 32.00 & 52.41 & 62.81 & 56.41
 \\
\ourdino{} ViT-2B & 45.99 & 1666.01 & 60.13 & 66.64 & 60.19 & 68.71 & 34.88 & 12.10 & 62.07 & 18.60 & 4.39 & 45.55 & 14.00 & 32.66 & 52.67 & 62.07 & 57.83 \\
\ourdino{} ViT-3B & 46.43 & 1729.40 & 60.56 & 66.99 & 60.24 & 70.50 & 31.88 & 11.70 & 62.30 & 17.52 & 4.80 & 45.18 & 15.00 & 31.33 & 53.20 & 62.77 & 62.50 \\
\ourdino{} ViT-5B & 46.28 & 1661.25 & 59.27 & 67.24 & 61.10 & 69.41 & 31.55 & 10.90 & 61.46 & 18.72 & 4.60 & 45.53 & 15.00 & 34.00 & 53.07 & 64.57 & 61.08 \\

    \end{tabular}
\end{adjustbox}
\caption{\textbf{VQA Evaluation: \ourdino{} trained on ImageNet-1k.}}
\label{tab:VQA PIN-DINO ImageNet Data-scale}
\end{table*}

\subsection{\ourmae{}}

We show VQA evaluation results from scaling up MAE trained on \ourdata{}, in \cref{tab:VQA PIN-MAE MC-2B}. These are the full results for Question~\ref{rq:generalize to other ssl}.

\begin{table*}[ht]
\centering
    \small  
    \setlength\tabcolsep{2pt} 
    \begin{adjustbox}{max width=\textwidth}
    \begin{tabular}{l|r|rrrr|rrrr|rrrr|rrrr}
     \multicolumn{1}{c|}{Vision Backbone} &  \multicolumn{1}{c|}{} & \multicolumn{4}{c|}{General} & \multicolumn{4}{c|}{Knowledge} & \multicolumn{4}{c|}{OCR \& Chart} & \multicolumn{4}{c}{Vision-Centric}  \\
      Model  & \multicolumn{1}{c|}{\rotatebox{90}{Average}} & \multicolumn{1}{c}{\rotatebox{90}{MME$^\text{P}$}} & \multicolumn{1}{c}{\rotatebox{90}{MMB}} & \multicolumn{1}{c}{\rotatebox{90}{SEED$^\text{I}$}} & \multicolumn{1}{c|}{\rotatebox{90}{GQA}} & \multicolumn{1}{c}{\rotatebox{90}{SQA$^\text{I}$}} & \multicolumn{1}{c}{\rotatebox{90}{MMMU$^\text{V}$}} & \multicolumn{1}{c}{\rotatebox{90}{MathVista$^\text{M}$}} & \multicolumn{1}{c|}{\rotatebox{90}{AI2D}} & \multicolumn{1}{c}{\rotatebox{90}{ChartQA}} & \multicolumn{1}{c}{\rotatebox{90}{OCRBench}} & \multicolumn{1}{c}{\rotatebox{90}{TextVQA}} & \multicolumn{1}{c|}{\rotatebox{90}{DocVQA}} & \multicolumn{1}{c}{\rotatebox{90}{MMVP}} & \multicolumn{1}{c}{\rotatebox{90}{RealWorldQA}} & \multicolumn{1}{c}{\rotatebox{90}{CV-Bench$^\text{2D}$}} & \multicolumn{1}{c}{\rotatebox{90}{CV-Bench$^\text{3D}$}} \\
\hline 
\ourmae{} ViT-1B & 49.19 & 1736.22 & 62.02 & 68.38 & 60.05 & 73.27 & 33.11 & 12.90 & 63.92 & 23.60 & 16.40 & 47.84 & 18.00 & 36.66 & 52.81 & 70.42 & 60.83

 \\
\ourmae{} ViT-2B & 50.59 & 1700.16 & 63.57 & 69.21 & 60.93 & 72.48 & 32.22 & 15.50 & 64.44 & 29.00 & 23.20 & 48.78 & 20.00 & 38.00 & 55.16 & 67.98 & 63.91
\\
\ourmae{} ViT-3B & 50.92 & 1723.85 & 64.69 & 69.71 & 60.94 & 72.13 & 34.33 & 13.50 & 65.70 & 30.92 & 24.60 & 48.92 & 20.00 & 37.33 & 54.64 & 64.15 & 66.91
 \\
\ourmae{} ViT-5B & 51.50 & 1710.13 & 65.12 & 70.13 & 61.10 & 72.63 & 32.66 & 13.90 & 65.67 & 33.80 & 26.50 & 49.60 & 21.00 & 38.00 & 53.72 & 66.69 & 67.91
 \\

    \end{tabular}
\end{adjustbox}
\caption{\textbf{VQA Evaluation: \ourmae{} trained on \ourdata{}.}}
\label{tab:VQA PIN-MAE MC-2B}
\end{table*}

\subsection{Scaled CLIP Models}

We show VQA evaluation results from scaling up  MetaCLIP~\citep{xu2023demystifying} trained on \ourdata{}, in \cref{tab:VQA CLIP MC-2B}. These are the full results for \cref{subsec:scale vit}. In contrast to visual SSL methods in \cref{tab:Vision PIN-DINO MC-2B Data-2B} and \cref{tab:VQA PIN-MAE MC-2B}, CLIP models do not exhibit clear scaling behavior.

\begin{table*}[ht]
\centering
    \small  
    \setlength\tabcolsep{2pt} 
    \begin{adjustbox}{max width=\textwidth}
    \begin{tabular}{l|r|rrrr|rrrr|rrrr|rrrr}
     \multicolumn{1}{c|}{Vision Backbone} &  \multicolumn{1}{c|}{} & \multicolumn{4}{c|}{General} & \multicolumn{4}{c|}{Knowledge} & \multicolumn{4}{c|}{OCR \& Chart} & \multicolumn{4}{c}{Vision-Centric}  \\
      Model  & \multicolumn{1}{c|}{\rotatebox{90}{Average}} & \multicolumn{1}{c}{\rotatebox{90}{MME$^\text{P}$}} & \multicolumn{1}{c}{\rotatebox{90}{MMB}} & \multicolumn{1}{c}{\rotatebox{90}{SEED$^\text{I}$}} & \multicolumn{1}{c|}{\rotatebox{90}{GQA}} & \multicolumn{1}{c}{\rotatebox{90}{SQA$^\text{I}$}} & \multicolumn{1}{c}{\rotatebox{90}{MMMU$^\text{V}$}} & \multicolumn{1}{c}{\rotatebox{90}{MathVista$^\text{M}$}} & \multicolumn{1}{c|}{\rotatebox{90}{AI2D}} & \multicolumn{1}{c}{\rotatebox{90}{ChartQA}} & \multicolumn{1}{c}{\rotatebox{90}{OCRBench}} & \multicolumn{1}{c}{\rotatebox{90}{TextVQA}} & \multicolumn{1}{c|}{\rotatebox{90}{DocVQA}} & \multicolumn{1}{c}{\rotatebox{90}{MMVP}} & \multicolumn{1}{c}{\rotatebox{90}{RealWorldQA}} & \multicolumn{1}{c}{\rotatebox{90}{CV-Bench$^\text{2D}$}} & \multicolumn{1}{c}{\rotatebox{90}{CV-Bench$^\text{3D}$}} \\
\hline 
MetaCLIP ViT-1B & 52.30 & 1813.70 & 68.90 & 69.45 & 60.35 & 74.07 & 33.55 & 12.70 & 64.41 & 33.20 & 34.59 & 52.15 & 26.00 & 37.33 & 52.15 & 65.47 & 61.83
 \\
MetaCLIP ViT-2B & 53.03 & 1787.39 & 68.81 & 69.54 & 61.08 & 75.16 & 34.66 & 20.10 & 65.38 & 32.80 & 32.90 & 52.55 & 26.00 & 37.33 & 52.94 & 65.19 & 64.67
\\
MetaCLIP ViT-3B & 53.22 & 1873.67 & 68.72 & 70.33 & 61.85 & 77.29 & 32.77 & 11.80 & 66.35 & 32.16 & 34.40 & 54.58 & 26.00 & 35.33 & 55.55 & 65.57 & 65.08
 \\
MetaCLIP ViT-5B & 52.52 & 1779.03 & 70.10 & 70.26 & 61.53 & 72.43 & 33.44 & 17.90 & 66.74 & 30.04 & 32.20 & 52.49 & 25.00 & 39.33 & 54.50 & 64.22 & 61.16
 \\
MetaCLIP ViT-7B & 52.97 & 1827.80 & 69.93 & 69.47 & 61.33 & 74.91 & 35.55 & 16.80 & 65.15 & 32.12 & 32.10 & 52.07 & 25.00 & 39.33 & 54.11 & 65.08 & 63.16

 \\

    \end{tabular}
\end{adjustbox}
\caption{\textbf{VQA Evaluation: MetaCLIP trained on \ourdata{} with 2 billion images seen.}}
\label{tab:VQA CLIP MC-2B}
\end{table*}

\subsection{Text Filtered Models}
We provide full results for Question~\ref{rq:probe into text}. As shown in \cref{tab:VQA MC-DINO Text Filterd Models}, SSL models learn features particularly well-suited for OCR \& Chart tasks when trained on datasets with a higher concentration of text-rich images. This suggests that visual SSL is sensitive to the underlying training distribution and can be effectively steered toward specific downstream applications, such as OCR \& Chart.
\begin{table*}[ht]
\centering
    \small  
    \setlength\tabcolsep{2pt} 
    \begin{adjustbox}{max width=\textwidth}
    \begin{tabular}{l|r|rrrr|rrrr|rrrr|rrrr}
     \multicolumn{1}{c|}{Vision Backbone} &  \multicolumn{1}{c|}{} & \multicolumn{4}{c|}{General} & \multicolumn{4}{c|}{Knowledge} & \multicolumn{4}{c|}{OCR \& Chart} & \multicolumn{4}{c}{Vision-Centric}  \\
      Model  & \multicolumn{1}{c|}{\rotatebox{90}{Average}} & \multicolumn{1}{c}{\rotatebox{90}{MME$^\text{P}$}} & \multicolumn{1}{c}{\rotatebox{90}{MMB}} & \multicolumn{1}{c}{\rotatebox{90}{SEED$^\text{I}$}} & \multicolumn{1}{c|}{\rotatebox{90}{GQA}} & \multicolumn{1}{c}{\rotatebox{90}{SQA$^\text{I}$}} & \multicolumn{1}{c}{\rotatebox{90}{MMMU$^\text{V}$}} & \multicolumn{1}{c}{\rotatebox{90}{MathVista$^\text{M}$}} & \multicolumn{1}{c|}{\rotatebox{90}{AI2D}} & \multicolumn{1}{c}{\rotatebox{90}{ChartQA}} & \multicolumn{1}{c}{\rotatebox{90}{OCRBench}} & \multicolumn{1}{c}{\rotatebox{90}{TextVQA}} & \multicolumn{1}{c|}{\rotatebox{90}{DocVQA}} & \multicolumn{1}{c}{\rotatebox{90}{MMVP}} & \multicolumn{1}{c}{\rotatebox{90}{RealWorldQA}} & \multicolumn{1}{c}{\rotatebox{90}{CV-Bench$^\text{2D}$}} & \multicolumn{1}{c}{\rotatebox{90}{CV-Bench$^\text{3D}$}} \\
\hline 
\textcolor{gray}{\ourdino{} ViT-1B (No Filter)} & \textcolor{gray}{49.01} & \textcolor{gray}{1731.52} & \textcolor{gray}{65.37} & \textcolor{gray}{69.92} & \textcolor{gray}{62.40} & \textcolor{gray}{72.58} & \textcolor{gray}{35.33} & \textcolor{gray}{12.30} & \textcolor{gray}{64.28} & \textcolor{gray}{19.20} & \textcolor{gray}{9.40} & \textcolor{gray}{47.41} & \textcolor{gray}{17.00} & \textcolor{gray}{37.33} & \textcolor{gray}{57.12} & \textcolor{gray}{64.80} & \textcolor{gray}{63.16} \\

\ourdino{} ViT-1B (Light Filter) & 50.73 & 1690.89 & 65.54 & 70.68 & 62.63 & 70.99 & 33.89 & 17.80 & 63.69 & 26.12 & 21.80 & 50.56 & 20.00 & 36.00 & 56.86 & 64.84 & 65.75 \\

\ourdino{} ViT-1B (Heavy Filter) & 49.44 & 1593.79 & 61.40 & 65.34 & 59.53 & 71.19 & 31.33 & 14.90 & 64.83 & 36.92 & 24.09 & 50.09 & 27.00 & 21.33 & 53.20 & 66.53 & 63.66
 \\
\textcolor{gray}{\ourdino{} ViT-2B (No Filter)} & \textcolor{gray}{50.77} & \textcolor{gray}{1760.80} & \textcolor{gray}{68.98} & \textcolor{gray}{71.29} & \textcolor{gray}{62.89} & \textcolor{gray}{73.67} & \textcolor{gray}{31.77} & \textcolor{gray}{15.90} & \textcolor{gray}{67.06} & \textcolor{gray}{23.30} & \textcolor{gray}{15.60} & \textcolor{gray}{49.20} & \textcolor{gray}{19.00} & \textcolor{gray}{38.00} & \textcolor{gray}{57.38} & \textcolor{gray}{65.85} & \textcolor{gray}{64.41} \\
\ourdino{} ViT-2B (Light Filter) & 53.38 & 1768.67 & 68.38 & 71.80 & 63.24 & 74.16 & 33.88 & 31.40 & 67.38 & 31.40 & 27.30 & 51.26 & 23.00 & 39.33 & 56.47 & 61.13 & 65.50

 \\
\ourdino{} ViT-2B (Heavy Filter) & 53.65 & 1743.56 & 65.29 & 69.28 & 61.19 & 74.86 & 32.22 & 14.50 & 67.42 & 47.48 & 29.40 & 52.80 & 32.00 & 40.00 & 54.50 & 65.85 & 64.50

    \end{tabular}
\end{adjustbox}
\caption{\textbf{VQA Evaluation: \ourdino{} trained on text filtered \ourdata{}.}}
\label{tab:VQA MC-DINO Text Filterd Models}
\end{table*}

\subsection{Baseline Models}
In \cref{tab:VQA Reference Models}, we provide full VQA results for the reference off-shelf models that we evaluated in \cref{sec:final model}. 

\begin{table*}[ht]
\centering
    \small  
    \setlength\tabcolsep{2pt} 
    \begin{adjustbox}{max width=\textwidth}
    \begin{tabular}{l|r|rrrr|rrrr|rrrr|rrrr}
     \multicolumn{1}{c|}{Vision Backbone} &  \multicolumn{1}{c|}{} & \multicolumn{4}{c|}{General} & \multicolumn{4}{c|}{Knowledge} & \multicolumn{4}{c|}{OCR \& Chart} & \multicolumn{4}{c}{Vision-Centric}  \\
      Model  & \multicolumn{1}{c|}{\rotatebox{90}{Average}} & \multicolumn{1}{c}{\rotatebox{90}{MME$^\text{P}$}} & \multicolumn{1}{c}{\rotatebox{90}{MMB}} & \multicolumn{1}{c}{\rotatebox{90}{SEED$^\text{I}$}} & \multicolumn{1}{c|}{\rotatebox{90}{GQA}} & \multicolumn{1}{c}{\rotatebox{90}{SQA$^\text{I}$}} & \multicolumn{1}{c}{\rotatebox{90}{MMMU$^\text{V}$}} & \multicolumn{1}{c}{\rotatebox{90}{MathVista$^\text{M}$}} & \multicolumn{1}{c|}{\rotatebox{90}{AI2D}} & \multicolumn{1}{c}{\rotatebox{90}{ChartQA}} & \multicolumn{1}{c}{\rotatebox{90}{OCRBench}} & \multicolumn{1}{c}{\rotatebox{90}{TextVQA}} & \multicolumn{1}{c|}{\rotatebox{90}{DocVQA}} & \multicolumn{1}{c}{\rotatebox{90}{MMVP}} & \multicolumn{1}{c}{\rotatebox{90}{RealWorldQA}} & \multicolumn{1}{c}{\rotatebox{90}{CV-Bench$^\text{2D}$}} & \multicolumn{1}{c}{\rotatebox{90}{CV-Bench$^\text{3D}$}} \\
\hline 
\rowcolor{gray!10} CLIP Models    &  &  &  &  &  &  &   &  &  & &  &  & & &  &  & 
\\
MetaCLIP ViT-H$_{224\text{px}}$ & 54.91 & 1860.58 & 72.93 & 70.96 & 62.22 & 77.88 & 36.88 & 15.00 & 67.32 & 35.60 & 33.40 & 55.10 & 29.00 & 41.33 & 53.46 & 68.53 & 65.91 \\
SigLIP ViT-SO400M$_{224\text{px}}$ & 55.36 & 1807.30 & 72.76 & 71.83 & 62.68 & 76.74 & 35.44 & 14.00 & 68.65 & 33.08 & 40.20 & 56.61 & 28.00 & 47.33 & 56.99 & 66.42 & 64.66 
 \\
SigLIP ViT-SO400M$_{384\text{px}}$ & 59.97 & 1892.16 & 73.71 & 73.00 & 63.80 & 77.83 & 33.88 & 20.00 & 69.78 & 54.24 & 46.40 & 63.53 & 50.00 & 46.00 & 58.43 & 67.37 & 66.91

 \\

SigLIP2 ViT-SO400M$_{224\text{px}}$ & 56.32 & 1789.26 & 73.36 & 72.20 & 62.60 & 74.96 & 35.55 & 22.40 & 69.85 & 35.76 & 42.00 & 59.68 & 31.00 & 44.00 & 54.24 & 69.88 & 64.16
\\

SigLIP2 ViT-SO400M$_{384\text{px}}$ & 61.98 & 1895.70 & 74.57 & 72.24 & 64.81 & 79.27 & 36.33 & 19.90 & 72.24 & 59.68 & 52.90 & 67.15 & 54.00 & 49.33 & 54.77 & 70.73 & 69.00

\\

\rowcolor{gray!10}SSL Models    &  &  &  &  &  &  &   &  &  & &  &  & & &  &  & 
\\
DINOv2 ViT-g$_{224\text{px}}$  & 49.25 & 1785.25 & 64.86 & 70.89 & 62.89 & 72.03 & 32.11 & 12.40 & 62.37 & 17.96 & 5.50 & 47.06 & 15.00 & 47.33 & 56.33 & 65.92 & 66.08
\\

DINOv2 ViT-g$_{378\text{px}}$ & 47.94 & 1734.38 & 64.26 & 71.50 & 62.21 & 71.04 & 33.11 & 9.60 & 63.08 & 17.76 & 5.00 & 45.59 & 15.00 & 41.33 & 56.47 & 63.79 & 60.58

\\

DINOv2 ViT-g$_{518\text{px}}$ & 47.91 & 1694.08 & 62.45 & 70.64 & 62.87 & 71.29 & 33.55 & 11.80 & 63.37 & 18.32 & 5.10 & 46.27 & 15.00 & 37.33 & 56.60 & 65.36 & 61.83

\\
I-JEPA ViT-H $_{224\text{px}}$ & 44.78 & 1598.15 & 60.01 & 64.04 & 57.66 & 68.91 & 34.55 & 10.20 & 62.07 & 16.72 & 4.00 & 42.99 & 14.00 & 29.33 & 49.93 & 57.39 & 57.16

 \\
MAE ViT-H$_{224\text{px}}$ & 45.21 & 1697.06 & 56.87 & 56.41 & 60.51 & 70.74 & 32.11 & 11.50 & 61.30 & 17.40 & 5.50 & 45.38 & 14.00 & 27.33 & 53.46 & 61.19 & 64.75

 \\

    \end{tabular}
\end{adjustbox}
\caption{\textbf{VQA Evaluation: Off-shelf CLIP and SSL models.}}
\label{tab:VQA Reference Models}
\end{table*}

\section{High Resolution Adaption of \ourssl{}}
\label{appendix:high res}
Following \citet{oquab2023dinov2}, we further fine-tune our model under higher resolution settings of 378$\times$378 and 518$\times$518 for 20k iterations. We use a batch size of 2048 and a correspondingly lower learning rate of 1.41e-5. All other parameters remain exactly the same as previously specified, including the learning rate warmup ratio, given the total of 10k iterations.

We also provided detailed benchmark results of high-resolution adaptation of \ourdino{} in \cref{tab:Our Highres Models}.

\begin{table*}[ht]
\centering
    \small  
    \setlength\tabcolsep{2pt} 
    \begin{adjustbox}{max width=\textwidth}
    \begin{tabular}{l|r|rrrr|rrrr|rrrr|rrrr}
     \multicolumn{1}{c|}{Vision Backbone} &  \multicolumn{1}{c|}{} & \multicolumn{4}{c|}{General} & \multicolumn{4}{c|}{Knowledge} & \multicolumn{4}{c|}{OCR \& Chart} & \multicolumn{4}{c}{Vision-Centric}  \\
      Model  & \multicolumn{1}{c|}{\rotatebox{90}{Average}} & \multicolumn{1}{c}{\rotatebox{90}{MME$^\text{P}$}} & \multicolumn{1}{c}{\rotatebox{90}{MMB}} & \multicolumn{1}{c}{\rotatebox{90}{SEED$^\text{I}$}} & \multicolumn{1}{c|}{\rotatebox{90}{GQA}} & \multicolumn{1}{c}{\rotatebox{90}{SQA$^\text{I}$}} & \multicolumn{1}{c}{\rotatebox{90}{MMMU$^\text{V}$}} & \multicolumn{1}{c}{\rotatebox{90}{MathVista$^\text{M}$}} & \multicolumn{1}{c|}{\rotatebox{90}{AI2D}} & \multicolumn{1}{c}{\rotatebox{90}{ChartQA}} & \multicolumn{1}{c}{\rotatebox{90}{OCRBench}} & \multicolumn{1}{c}{\rotatebox{90}{TextVQA}} & \multicolumn{1}{c|}{\rotatebox{90}{DocVQA}} & \multicolumn{1}{c}{\rotatebox{90}{MMVP}} & \multicolumn{1}{c}{\rotatebox{90}{RealWorldQA}} & \multicolumn{1}{c}{\rotatebox{90}{CV-Bench$^\text{2D}$}} & \multicolumn{1}{c}{\rotatebox{90}{CV-Bench$^\text{3D}$}} \\
\hline

\ourdino{}$_{224\text{px}}$ &  55.24 & 1811.05 & 71.30 & 72.14 & 64.04 & 72.43 & 35.66 & 15.20 & 68.52 & 35.52 & 36.40 & 56.53 & 29.00 & 46.00 & 57.90 & 70.53 & 62.08

\\
\ourdino{}$_{378\text{px}}$ & 57.43 & 1757.06 & 70.61 & 72.59 & 64.50 & 72.53 & 35.11 & 16.10 & 67.09 & 52.04 & 42.19 & 61.51 & 46.00 & 38.00 & 59.08 & 66.55 & 67.16

\\
\ourdino{}$_{518\text{px}}$ &  59.91 & 1807.08 & 73.79 & 72.92 & 64.78 & 74.36 & 34.66 & 14.50 & 69.43 & 57.28 & 45.70 & 64.48 & 53.00 & 43.33 & 60.52 & 70.08 & 69.41

    \end{tabular}
\end{adjustbox}
\caption{\textbf{VQA Evaluation: \ourdino{} ViT-7B adapted to different resolution}}
\label{tab:Our Highres Models}
\end{table*}

\section{Evaluation}

\cref{tab:benchmarks} lists evaluation benchmarks used and their purposes.

\begin{table}[htbp]
\centering
\begin{tabular}{lll}
\hline
Benchmark & Eval & Citation \\
\hline
GQA & General VQA & \citet{hudson2019gqa} \\
SEED & General VQA & \citet{ge2023planting} \\
MME & General VQA & \citet{fu2023mme} \\
MMBench & General VQA & \citet{liu2023mmbench} \\

AI2D & Knowledge VQA & \citet{hiippala2021ai2d} \\
ScienceQA & Knowledge VQA & \citet{lu2022learn} \\
MathVista & Knowledge VQA & \citet{lu2023mathvista} \\
MMMU & Knowledge VQA & \citet{yue2023mmmu} \\

TextVQA & OCR \& Chart VQA & \citet{singh2019towards} \\
DocVQA & OCR \& Chart VQA & \citet{mathew2021docvqa} \\
ChartQA & OCR \& Chart VQA & \citet{masry2022chartqa} \\
OCRBench & OCR \& Chart VQA & \citet{liu2023hidden} \\

MMVP & Vision-Centric VQA & \citet{tong2024eyes} \\
RealWorldQA & Vision-Centric VQA & \citet{grok} \\
CVBench-2D & Vision-Centric VQA & \citet{tong2024cambrian} \\
CVBench-3D & Vision-Centric VQA & \citet{tong2024cambrian} \\
ImageNet-1k & Image Classification & \citet{deng2009imagenet} \\
ADE-20k & Image Segmentation &\citet{zhou2019semantic} \\
NYU Depth v2 & Depth Estimation & \citet{silberman2012indoor} \\

\hline
\end{tabular}
\caption{\textbf{List of benchmarks used}}
\label{tab:benchmarks}
\end{table}

\section{Pretraining Dataset Cards}
\label{appendix:pretrain_dataset}
For reference, in \cref{tab:lvd_data_statistics} we include the data composition of LVD-142M, which was used to train the off-shelf DINOv2 model~\citep{oquab2023dinov2}. LVD-142M is a carefully curated data mix closely aligned with downstream classic vision evaluation tasks.
In comparison, we leverage MetaCLIP data, which is less curated and collected from 15 snapshots of CommonCrawl (CC).

\begin{table*}
\centering
\footnotesize
\begin{tabular}{lllrrr}
\hline
Task & Dataset / Split & Images & Retrieval & Retrieved & Final \\
\hline
classification & ImageNet-22k / -- & 14,197,086 & as is & -- & 14,197,086 \\
classification & ImageNet-22k / -- & 14,197,086 & sample & 56,788,344 & 56,788,344 \\
classification & ImageNet-1k / train & 1,281,167 & sample & 40,997,344 & 40,997,344 \\
\hline
fine-grained classif. & Caltech 101 / train & 3,030 & cluster & 2,630,000 & 1,000,000 \\
fine-grained classif. & CUB-200-2011 / train & 5,994 & cluster & 1,300,000 & 1,000,000 \\
fine-grained classif. & DTD / train1 & 1,880 & cluster & 1,580,000 & 1,000,000 \\
fine-grained classif. & FGVC-Aircraft / train & 3,334 & cluster & 1,170,000 & 1,000,000 \\
fine-grained classif. & Flowers-102 / train & 1,020 & cluster & 1,060,000 & 1,000,000 \\
fine-grained classif. & Food-101 / train & 75,750 & cluster & 21,670,000 & 1,000,000 \\
fine-grained classif. & Oxford-IIIT Pet / trainval & 3,680 & cluster & 2,750,000 & 1,000,000 \\
fine-grained classif. & Stanford Cars / train & 8,144 & cluster & 7,220,000 & 1,000,000 \\
fine-grained classif. & SUN397 / train1 & 19,850 & cluster & 18,950,000 & 1,000,000 \\
fine-grained classif. & Pascal VOC 2007 / train & 2,501 & cluster & 1,010,000 & 1,000,000 \\
\hline
segmentation & ADE20K / train & 20,210 & cluster & 20,720,000 & 1,000,000 \\
segmentation & Cityscapes / train & 2,975 & cluster & 1,390,000 & 1,000,000 \\
segmentation & Pascal VOC 2012 (seg.) / trainaug & 1,464 & cluster & 10,140,000 & 1,000,000 \\
\hline
depth estimation & Mapillary SLS / train & 1,434,262 & as is & -- & 1,434,262 \\
depth estimation & KITTI / train (Eigen) & 23,158 & cluster & 3,700,000 & 1,000,000 \\
depth estimation & NYU Depth V2 / train & 24,231 & cluster & 10,850,000 & 1,000,000 \\
depth estimation & SUN RGB-D / train & 4,829 & cluster & 4,870,000 & 1,000,000 \\
\hline
retrieval & Google Landmarks v2 / train (clean) & 1,580,470 & as is & -- & 1,580,470 \\
retrieval & Google Landmarks v2 / train (clean) & 1,580,470 & sample & 6,321,880 & 6,321,880 \\
retrieval & AmsterTime / new & 1,231 & cluster & 960,000 & 960,000 \\
retrieval & AmsterTime / old & 1,231 & cluster & 830,000 & 830,000 \\
retrieval & Met / train & 397,121 & cluster & 62,860,000 & 1,000,000 \\
retrieval & Revisiting Oxford / base & 4,993 & cluster & 3,680,000 & 1,000,000 \\
retrieval & Revisiting Paris / base & 6,322 & cluster & 3,660,000 & 1,000,000 \\
\hline
&&&&&142,109,386 \\
\hline
\end{tabular}
\caption{\textbf{LVD-142M Data Sources.} In contrast to LVD-142M, which relies on highly curated data sources drawn from distributions closely aligned with various downstream evaluation tasks (see the table above from \citet{oquab2023dinov2}), our data curation approach adopts the methodology from MetaCLIP~\citep{xu2023demystifying}, utilizing web data collected from 15 snapshots of CommonCrawl (CC) spanning January 2021 through January 2023.}
\label{tab:lvd_data_statistics}

\end{table*}